\documentclass[journal]{IEEEtran} %Ë«À¸
\usepackage{textcomp}
\usepackage{booktabs}
\usepackage{colortbl}
\usepackage[table]{xcolor}
\usepackage{graphicx}
\usepackage{subfigure}
\usepackage{multirow}
\usepackage{longtable}
\usepackage{algorithm}
\usepackage{algorithmic}
\usepackage{booktabs}
\usepackage{stfloats}
\usepackage{caption2}
\usepackage{amssymb,amsmath}
\usepackage{epstopdf}
\usepackage{color}
\usepackage{rotating}
\usepackage{multirow}
\usepackage{threeparttable}
\usepackage{bbding}
\usepackage{tabularx}
\usepackage{booktabs}
\usepackage{cite}
\usepackage[colorlinks,linkcolor=red,anchorcolor=blue,citecolor=blue]{hyperref}
\definecolor{rrr}{rgb}{1    0	0.0392}
\definecolor{myred}{rgb}{0.7059    0	0.0392}					% revised contents for reviewer 1
\usepackage{colortbl} % 用于给表格的行列上色
\definecolor{tabletitle}{gray}{.8}
\definecolor{ours}{gray}{.95}
\definecolor{ggray}{RGB}{127,127,127}
\definecolor{reda}{RGB}{245,60,50}
\definecolor{redb}{RGB}{217,148,143}
\definecolor{myyellow}{RGB}{190,144,0}
\definecolor{mygreen}{RGB}{0,136,51}
\definecolor{myblue}{RGB}{0,102,204}
\newcolumntype{B}{!{\vrule width 1pt}}
\usepackage{pifont}
\usepackage{orcidlink}

\ifCLASSINFOpdf
\else
\fi
\usepackage{xcolor}
\definecolor{myred}{rgb}{0.7059, 0, 0.0392}

\usepackage[normalem]{ulem}

\usepackage[
    replacedcolor=myred,
]{changes}

\setaddedmarkup{\textcolor{myred}{#1}}            
\setdeletedmarkup{\textcolor{myred}{\sout{#1}}}   
\begin{document}
\title{ASGNet: Adaptive Spectrum Guidance Network for Automatic Polyp Segmentation}
\author{
        Yanguang~Sun$^{\orcidlink{0009-0006-3765-6646}}$, 
        Hengmin~Zhang,
        Jianjun~Qian,
	      Jian~Yang, %\emph{Member,IEEE},
        Lei~Luo %\textsuperscript{\Envelope}
\thanks{This work was supported by the National Natural Science Foundation of China (Grant No. 62276135, 61806094, and 62176124), National Natural Science Fund for Excellent Young Scientists Fund Program (Overseas), and in part by the Open Project Program of Shanghai Key Laboratory of Data Science (No.2025090600010).%Natural Science Research Project of Colleges and Universities in Anhui Province (Grant No. xxxxxxx). 
(\textit{Corresponding author: Lei Luo,} luoleipitt@gmail.com)
	
Y. Sun, L. Luo, J. Qian, and J. Yang are with the PCA Laboratory, Key Laboratory of Intelligent Perception and Systems for High-Dimensional Information of Ministry of Education, School of Computer Science and Engineering, Nanjing University of Science and Technology, Nanjing, China. 
%(e-mail: Sunyg@njust.edu.cn; luoleipitt@gmail.com; csjqian@njust.edu.cn; csjyang@mail.njust.edu.cn).

H. Zhang is with the School of Information Science and Engineering, East China University of Science and Technology, Shanghai, China.

%J. Yang is with the College of Computer Science and Engineering, Nankai University, Tianjin, 300000, China. (csjyang@nankai.edu.cn)
}
}

\markboth{Journal of \LaTeX\ Class Files}%
{Shell \MakeLowercase{\textit{et al.}}: Bare Demo of IEEEtran.cls for IEEE Journals}

\maketitle
\begin{abstract}
Early identification and removal of polyps can reduce the risk of developing colorectal cancer. However, the diverse morphologies, complex backgrounds and often concealed nature of polyps make polyp segmentation in colonoscopy images highly challenging. Despite the promising performance of existing deep learning-based polyp segmentation methods, their perceptual capabilities remain biased toward local regions, mainly because of the strong spatial correlations between neighboring pixels in the spatial domain. This limitation makes it difficult to capture the complete polyp structures, ultimately leading to sub-optimal segmentation results. In this paper, we propose a novel adaptive spectrum guidance network, called ASGNet, which addresses the limitations of spatial perception by integrating spectral features with global attributes. Specifically, we first design a spectrum-guided non-local perception module that jointly aggregates local and global information, therefore enhancing the discriminability of polyp structures, and refining their boundaries. Moreover, we introduce a multi-source semantic extractor that integrates rich high-level semantic information to assist in the preliminary localization of polyps. Furthermore, we construct a dense cross-layer interaction decoder that effectively integrates diverse information from different layers and strengthens it to generate high-quality representations for accurate polyp segmentation. Extensive quantitative and qualitative results demonstrate the superiority of our ASGNet approach over 21 state-of-the-art methods across five widely-used polyp segmentation benchmarks. The code will be publicly available at: \href{https://github.com/CSYSI/ASGNet}{\color{blue} https://github.com/CSYSI/ASGNet}.
\end{abstract}
\IEEEpeerreviewmaketitle

\section{INTRODUCTION}
\label{S1}
\IEEEPARstart{C}{olorectal} cancer (CRC) is a common malignant tumor that occurs primarily in the colon and rectum, and it {ranks among the most} fatal cancers worldwide. Over time, CRC incidence and mortality rates have declined to some extent, but it remains a serious threat to human health, imposing significant psychological, economic, and social burdens on patients and their families \cite{C1,C3}. Most colorectal cancers develop from polyps in the colon or rectum, so early detection and removal of these polyps are keys to preventing CRC. However, accurate identification of polyps remains challenging due to various factors \cite{C6}. In particular, some polyps closely mimic the surrounding mucosa in color and texture, obscuring their boundaries. Furthermore, special types of polyps ($e.g.$, small or flat polyps) may elude detection during colonoscopy, especially when they are hidden within the intricate folds or bends of the colonic mucosa. Considering these factors, it is essential to develop computer-aided polyp segmentation methods to assist physicians in promptly detecting early-stage polyps, improving the accuracy of polyp identification.

\begin{figure}
   	   	\centering\includegraphics[width=0.48\textwidth,height=4.1cm]{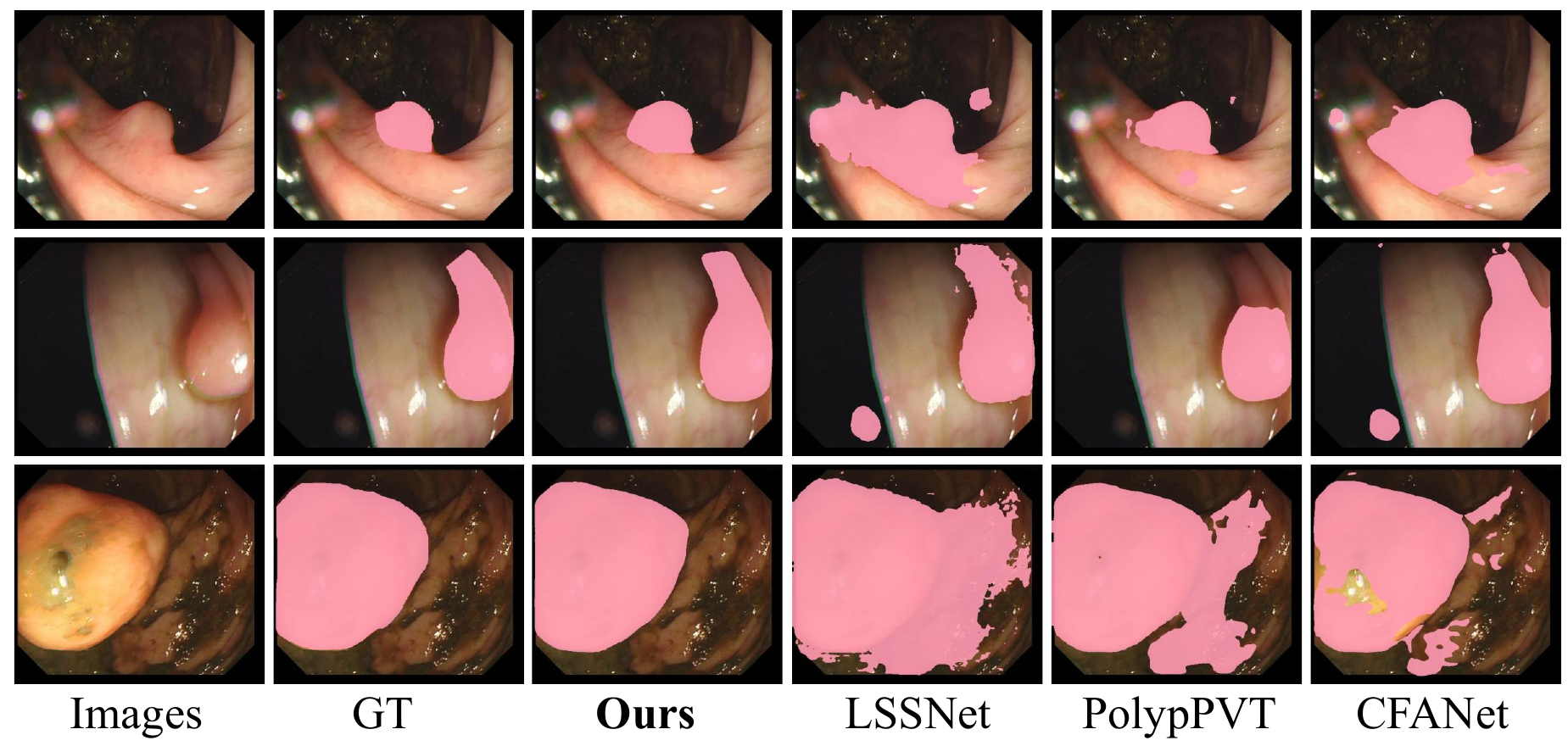}
   	   	\caption{Visualization results predicted by existing models ($i.e.$, CFANet \cite{CFANet}, PolypPVT \cite{PolypPVT}, and LSSNet \cite{LSSNet}) and ASGNet.}
   	   	\label{com_v1}		
\end{figure}

Early polyp segmentation methods \cite{T1,texture1} usually relied on hand-crafted features ($e.g.$, shape \cite{shape1} and texture \cite{texture1}) in traditional strategies. Due to their limited semantic understanding of image content, these methods often underperform when identifying challenging polyps. With the rapid development of deep learning, convolutional neural networks (CNNs) have been widely researched, enabling them to extract semantic features from input images. Consequently, numerous CNNs-based models \cite{UNet,UNet++,SANet} have been proposed for accurate polyp segmentation tasks. For example, SANet \cite{SANet} developed a shallow attention module to remove background noise from convolutional features. CFANet \cite{CFANet} integrated features of adjacent encoding layers to improve segmentation accuracy. Similarly, other methods have achieved impressive results by employing various optimization strategies, such as boundary guidance \cite{SFA,TCSVT-b,Polyper}, semantic enhancement \cite{Semantic_en}, and multi-scale learning \cite{MEGANet,TCSVT2,GLCONet,DMINet}. Although convolution-based methods have outperformed traditional models \cite{T1,texture1}, their limited receptive fields lead them to focus on local regions, making it difficult to capture global relationships. This may lead to detection errors or incomplete segmentation of polyps, as depicted by the CFANet \cite{CFANet} model in Fig. \ref{com_v1}.

To overcome this limitation, some polyp segmentation models \cite{PolypPVT, PPNet, LSSNet, DPU-Former} based on Transformer architectures have been designed to capture long-range dependencies across all pixels through self-attention mechanisms, thus improving global perception. Specifically, PolypPVT \cite{PolypPVT} adopted the pyramid vision transformer \cite{PVTv2} as an encoder to obtain multi-level features with global properties. MSBP \cite{MSBP} adopted a multi-scale boundary learning to optimize Transformer features. Despite achieving excellent performance, these methods consider only global relations between all pixels in the encoder and neglect the equally important decoder component in polyp segmentation, which is a sub-optimal strategy (as shown in PolypPVT \cite{PolypPVT} and LSSNet \cite{LSSNet} in Fig. \ref{com_v1}). In addition, adjacent pixels in the spatial domain typically exhibit strong correlations. This inherent local structure tends to bias the self-attention weights toward nearby regions, which to some extent weakens global modeling. Recently, spectrum information obtained through frequency transforms has been shown to possess global characteristics and has proven effective in enhancing the holistic understanding of images \cite{Fda, TCSVT6_FP, FSEL, GFNet, UDCNet}. This helps overcome the limitations of spatial features. To this end, several methods \cite{FTPS1, FTPS2, PSTNet} based on frequency transforms have been proposed to enhance initial features by integrating high-frequency and low-frequency components. 

Unlike these approaches, we re-examine and fully exploit the advantages of spectral features. On the one hand, we no longer simply divide spectral features into high and low frequencies \cite{FTPS2,FTPS1}, as this strategy tends to overlook significant information in the mid-frequency range. Instead, we apply the Fourier transform to convert pixel-level features into full-spectrum frequency features. On the other hand, rather than relying on simple filtering optimization \cite{FTPS1}, we enhance meaningful frequency bands within spectral features by introducing an adaptive weighting mechanism across both channel and spatial dimensions, while suppressing irrelevant background components. Furthermore, we extend the use of spectral features to the entire network architecture, rather than restricting them to a specific module, and use them to guide spatial features, enabling better focus on the structural information of polyps. Therefore, the polyps (as shown in ``Ours'' in Fig. \ref{com_v1}) predicted by our ASGNet are more accurate and complete through the guidance of spectrum information.

In this paper, we propose a novel adaptive spectrum guidance network (ASGNet), consisting of three key components: the spectrum-guided non-local perception module, {the} multi-source semantic extractor, and the dense cross-layer interaction decoder, which together achieve accurate polyp segmentation by leveraging spectrum information guidance. Technically, we design a spectrum-guided non-local perception (SNP) module that optimizes and reconstructs initial features from both global and local perspectives, using a joint-domain global optimization block and a local enhancement block. Meanwhile, we introduce a multi-source semantic extractor (MSE) to obtain higher-level semantics with different receptive fields, which helps guide the rough localization of polyps in the bottom features. Furthermore, we utilize a dense cross-layer interaction (DCI) decoder to generate discriminative features that include complete polyp structures and sharp boundaries through a series of interaction and reinforcement strategies, enabling precise segmentation of polyps. Extensive results demonstrate that our ASGNet achieves superior performance, outperforming 21 state-of-the-art (SOTA) methods. 

\indent Our main contributions are summarized as follows:

$\bullet$ A novel adaptive spectrum guidance network is proposed to generate high-quality features through joint frequency and spatial optimization for accurate polyp segmentation tasks. 

$\bullet$ Spectrum-guided non-local perception module is designed to simultaneously capture local and global information, thereby enriching and enhancing the initial features.

$\bullet$ Multi-source semantic extractor is introduced to enhance the ability of low-level features to identify polyps by aggregating higher-level semantics from various receptive fields.

$\bullet$ Dense cross-layer interaction decoder is utilized to aggregate and strengthen complementary information from different feature layers for better inference of polyp objects.

%The remainder of this paper is organized as follows. \textbf{Section \ref{S2}} reviews the related works. \textbf{Section \ref{S3}} gives the details of our ASGNet model. Extensive experimental results are adequately discussed and analyzed in \textbf{Section \ref{S4}}. Finally, the conclusion is summarized in \textbf{Section \ref{S5}}.

\section{Related Work}
\label{S2}
\subsection{\textbf{Polyp Segmentation Methods}}
 {Early} polyp segmentation methods \cite{T1,texture1} mainly relied on low-level visual priors ($e.g.$, texture \cite{texture1}, and shape \cite{shape1}) to predict colon polyps. However, {these} traditional methods often yield inaccurate segmentation results for challenging polyps due to their lack of high-level semantic perception. With the continuous advancement of deep learning, numerous convolution-based polyp segmentation methods \cite{FCN1,TCSVT2,TCSVT4} have achieved superior performance. {In particular}, Brandao $et$ $al.$ \cite{FCN1} designed a CNNs-based framework to segment polyps and high-risk regions in colonoscopy images. Subsequently, Fang $et$ $al.$ \cite{SFA} utilized a shared encoder and two mutually constrained decoders to detect polyp regions. Zhou $et$ $al.$ \cite{CFANet} utilized simultaneously boundary and hierarchical semantic information to segment polyps. Furthermore, Dong $et$ $al.$ \cite{PolypPVT} and Hu $et$ $al.$ \cite{PPNet} exploited Transformer networks as encoders to extract initial features with long-range dependencies for segmenting polyps. He $et$ $al.$ \cite{RUN} proposed a reversible unfolding network to achieve concealed object segmentation.  Kamara $et$ $al.$ \cite{MDPNet} designed a multiscale dynamic polyp-focus network to solve the low contrast between polyps and the surrounding background. Pan $et$ $al.$ \cite{MSBP} designed a multi-scale boundary prediction network for effective polyp segmentation with low complexity. Although these methods \cite{EPSegNet, BFANet, TCNNF, CFANet} have achieved promising performance, they remain constrained by the inherent limitations of pixel-level spatial features, which ultimately hinders segmentation accuracy. Consequently, we propose the ASGNet method, which addresses the limitations of spatial features by introducing global spectral information to reconstruct the encoding features and employing a series of optimization strategies to further enhance object–background discrimination, achieving accurate polyp segmentation.
\begin{figure*}
  	\centering\includegraphics[width=0.98\textwidth,height=3.9cm]{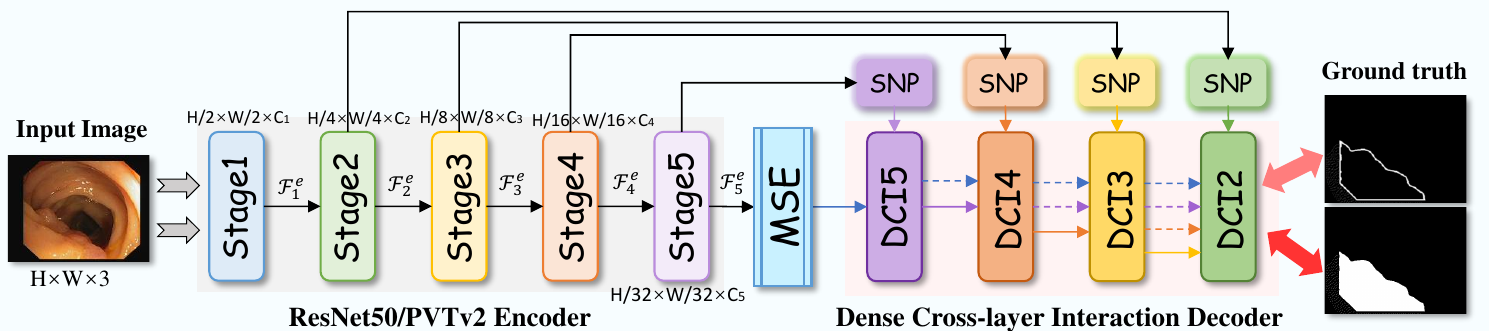}
  	\caption{Overall framework of the proposed ASGNet method, which consists of the {basic} encoder, the spectrum-guided non-local perception (SNP) module, the multi-source semantic extractor (MSE), and the dense cross-layer interaction (DCI) decoder.}
  	\label{ASGNet}		
\end{figure*}

\subsection{\textbf{Vision Transformers}}
Recently, Transformer-based models  have been widely adopted in computer vision, including tasks such as image classification \cite{ViT}, image editing \cite{yu2025ssaim, yu2025ttfdiffusion}, and object detection \cite{DSP, wang2025m4}, where their self-attention mechanisms enable the modeling of long-range dependencies, leading to impressive performance gains. Specifically, Yuan $et$ $al.$ \cite{ViT} were the first to employ the Transformer architecture to globally model and extract features from images. Wang $et$ $al.$ \cite{PVTv2} further proposed a feature pyramid structure to compensate for the lack of local-scale information. Li $et$ $al.$ \cite{UniFormer} integrated CNNs and self-attention within a unified block, combining the advantages of both architectures. Likewise, other Transformer-based models \cite{WEFT,Swin,Controllable-LPMoE,RCNet,wang2026localized,wang2025msod}, such as Swin Transformer \cite{Swin} and PARFormer \cite{TCSVT7_TF}, have also achieved remarkable performance. Different from them, we embed the ASF into both the self-attention and feed-forward network to further enhance global perception by aggregating spectrum information.

\subsection{\textbf{Spectrum Learning}}
Image analysis based on frequency-domain exploration has garnered significant attention. Some methods \cite{TCSVT6_FP,GFNet,FSEL,TCSVT5_FP,UDCNet} effectively transform local spatial features into global spectral features via frequency {transforms} to obtain diverse types of information, yielding impressive results. This is because frequency transforms project images onto globally supported basis functions, so each spectral coefficient aggregates information from the entire spatial domain and thus naturally encodes global properties and long-range dependencies. For example, Yang $et$ $al.$ \cite{Fda} adopted the Fourier transform to assist domain adaptation for boosting semantic segmentation. Rao $et$ $al.$ \cite{GFNet} proposed a global filter network to learn long-range dependencies in the frequency domain. Liu $et$ $al.$ \cite{FSI} exploited a frequency–spatial interactive learning network to incorporate spectrum learning into the image restoration. Wang $et$ $al.$ \cite{FFCNet} devised a frequency learning framework for colon disease classification. Zhu $et$ $al.$ \cite{FDTNet} designed a dual-stream frequency-aware detection network for prohibited object detection. In this paper, we design the ASF based on the Fast Fourier Transform, which guides features to focus on polyps by enhancing frequency features in specific bands.

\begin{figure*}
   	   	\centering\includegraphics[width=0.95\textwidth,height=6.6cm]{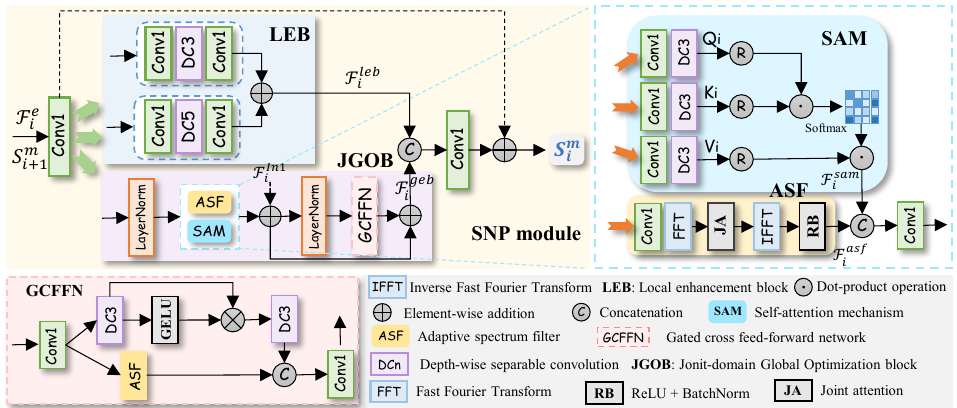}
   	   	\caption{Details of the spectrum-guided non-local perception module.}
   	   	\label{SNP_L}		
\end{figure*}
 
\section{Proposed ASGNet Method}
\label{S3}
\subsection{\textbf{Overall Architecture}}
\label{OA}
The overall architecture of our ASGNet is illustrated in Fig.~\ref{ASGNet}. It consists of four key components: (a) ResNet50 \cite{ResNet}/PVTv2 \cite{PVTv2} encoder, (b) Spectrum-guided Non-local Perception module, (c) Multi-source Semantic Extractor, and (d) Dense Cross-layer Interaction decoder. In particular, for the ResNet50 \cite{ResNet} encoder, an image $I_m\in\mathbb{R}^{H\times W\times3}$ is first fed to extract initial features$\{\mathcal{F} _{i}^{e}\}_{i=1}^{5}$ with a resolution of $\{\frac W{2^i},\frac H{2^i}\}$. {Since} low-level features contain considerable background noise and higher resolutions may slow down model training, we discard the feature $\mathcal{F}_{1}^{e}$ in ASGNet model and exploit only the features from the last four layers. Subsequently, initial features $\{\mathcal{F}_{i}^{e}\}_{i=2}^{5}$ are refined by the spectrum-guided non-local perception module to produce features $\{\mathcal{S} _{i}^{m}\}_{i=2}^{5}$ with abundant local-global information. Meanwhile, the multi-source semantic extractor captures diverse semantic by operating on the deep feature $\mathcal{F}_{5}^{e}$ to produce a higher-level feature $\mathcal{F}_{6}^{mse}$, which guides polyp localization. Finally, we employ the dense cross-layer interaction decoder to fuse complementary clues from the optimized features, generating high-quality representations $\{\mathcal{D} _{i}^{d}\}_{i=2}^{5}$ for polyp segmentation.

\subsection{\textbf{Spectrum-guided Non-local Perception Module}}
\label{SNPM}
Polyp segmentation in the intestinal environment is a highly challenging yet significant task that requires sufficient contextual information. {For that}, we propose the spectrum-guided non-local perception (SNP) module (as depicted in Fig. \ref{SNP_L}), which enhances the {discrimination} of initial features $\{\mathcal{F} _{i}^{e}\}_{i=2}^{5}$ for polyp regions by integrating semantic relationships and spatial details from the designed joint-domain global optimization block and local enhancement block. Specifically, \textbf{joint-domain global optimization block (JGOB)} employs both a self-attention mechanism (SAM) and an adaptive spectrum filter (ASF) to model relationships of all pixels. 

Technically, the initial feature $\mathcal{F} _{i}^{e}$ and the optimized feature $\mathcal{S} _{i+1}^{m}$ are first unified into {96} channels via a 1$\times$1 convolution ($\mathcal{C}_{1}$), $i.e.$, $\mathcal{F} _{i}^{In1}=\mathcal{C}_{1}[\mathcal{F} _{i}^{e}, \mathcal{S} _{i+1}^{m}]$, where $[\cdot]$ denotes the concatenation, and 3 $\le i+1 \le$ 5. Then, layer normalization is employed to improve model training stability. Next, we use a 1$\times$1 convolution ($\mathcal{C}_{1}$) to perform positional induction, and utilize a 3$\times$3 depthwise separable convolution ($\mathcal{DC}_3$) to obtain the $Key$ $(\mathcal{K}_i)$, $Query$ $(\mathcal{Q}_i)$, and $Value$ $(\mathcal{V}_i)$ required for the self-attention mechanism. The process is as follows:  
\begin{equation}
	\begin{split}
            & {[\mathcal{Q}_i,\thinspace\mathcal{K}_i,\thinspace\mathcal{V}_i]=\mathcal{DC}_3\mathcal{C}_1\mathbb{LN}(\mathcal{F}_{i}^{In1}), \ i=2,3,4,5} \\
	\end{split}
\end{equation}
where $\mathbb{LN}(\cdot)$ presents the layer normalization. Subsequently, we reshape $\mathcal{Q}_i$ and $\mathcal{K}_i$ and use the dot-product and Softmax operations to generate a global attention map $\mathcal{A}_i$, that is, $\mathcal{A}_i=\mathbb{SOF}(\overline{\mathcal{Q}_i}\odot\overline{\mathcal{K}_i}$), where $\odot$ denotes the dot-product operation, $\mathbb{SOF}(\cdot)$ is the Softmax,  $\overline{\mathcal{Q}_i}\in\mathbb{R}^{C\times HW}$ and $\overline{\mathcal{K}_i}\in\mathbb{R}^{HW\times C}$ are the reconstructed $\mathcal{Q}_i$ and $\mathcal{K}_i$, respectively. Afterward, we utilize {the} attention map $\mathcal{A}_i$ to model long-range dependencies between pixels to obtain the feature $\mathcal{F}_{i}^{sam}$, $i.e.$, $\mathcal{F}_{i}^{sam}=\mathcal{A}_i \odot\overline{\mathcal{V}_i}$. Although self-attention mechanisms enable models to learn long-range pixel dependencies, local correlations often remain the primary focus when dealing with spatial pixel features. Since each frequency coefficient is computed from the whole image, spectrum features naturally encode global context \cite{Fda,FSEL,TCSVT5_FP,GFNet}. Therefore, performing adaptive filtering in the frequency domain is equivalent to applying a global receptive field transformation in the spatial domain, which effectively enhances the model's global representation ability. Based on this, we propose an \textbf{adaptive spectrum filter (ASF)} to further enhance global representation ability by incorporating spectrum information of image-level receptive fields. Technically, we first apply the Fast Fourier Transform to map pixel features to spectral features with image-level receptive fields. Furthermore, we employ a joint attention to perform spatial and channel-wise adaptive calibration and optimization on these spectrum features, as follows:
\begin{equation}
	\begin{split}
            & \mathcal{F}_{i}^{In2}= \mathbb{JA}(\overrightarrow{\Psi}|\mathcal{C}_{1}\mathbb{LN}(\mathcal{F}_{i}^{In1})|),   \quad i= 2,3,4,5,\\
            & \mathbb{JA}(x)=\mathbb{MSA}(\mathbb{CA}(x)) \oplus x, \\
            & {\mathbb{CA}(x)=\sigma{_s}(\mathcal{C}_1(\mathbb{GAP}(x)+\mathbb{GMP}(x)))\otimes x},\\
            & {\mathbb{MSA}(x)=\sigma{_s}(\mathcal{C}_3\sigma{_r}(\mathcal{C}_3x))\otimes x \oplus \sigma{_s}(\mathcal{C}_5\sigma{_r}(\mathcal{C}_5x))\otimes x},\\
	\end{split}
\end{equation}
where $\mathbb{JA}(\cdot)$ represents the joint attention that contains a channel attention ($\mathbb{CA}(\cdot)$) and a multi-scale spatial attention ($\mathbb{MSA}(\cdot)$). $\overrightarrow{\Psi}|\cdot|$ denotes the Fast Fourier Transform, $i.e.$, ${\overrightarrow{\Psi}|\cdot| = \sum_{w=0}^{W-1}\sum_{h=0}^{H-1}x(w, h)e^{-2\pi i(\frac{uw}{W}+\frac{vh}{H})},}$ {$(w, h)$ and $(u, v)$ denote spatial domain coordinates and frequency domain coordinates, $i$ presents the imaginary part}, ``$\oplus$'' and ``$\otimes$'' are the element-wise addition and multiplication. ``$\sigma_s$'' and ``$\sigma_r$'' present the Sigmoid and ReLU activation functions. Afterwards, the optimized feature $\mathcal{F}_{i}^{In2}$ is transformed back to the original domain using the inverse Fast Fourier Transform. Subsequently, we adopt the modulus operation to obtain spectrum information in the frequency domain, and enhance its stability by applying the BatchNorm and the ReLU activation function, which can be formulated as:
\begin{equation}
	\begin{split}
            & \mathcal{F}_{i}^{asf}= \mathbb{RB}(\Xi (\overleftarrow{\Psi}|\mathcal{F}_{i}^{In2}|)),   \quad i= 2,3,4,5,\\
	\end{split}
\end{equation}
where $\mathbb{RB}(\cdot)$ denotes the BatchNorm and the ReLU function, $\Xi(\cdot)$ {presents} the modulus operation, and $\overleftarrow{\Psi}|\cdot|$ {represents} the inverse Fast Fourier Transform, {that is}, {$\overleftarrow{\Psi}|\cdot| =\frac{1}{WH}\sum_{u=0}^{W-1}\sum_{v=0}^{H-1}x(u,v)e^{2\pi i(\frac{uw}{W}+\frac{vh}{H} )}$}. Furthermore, we aggregate the optimized features to generate feature $\mathcal{F}_{i}^{In3}$ with {rich} global {relationships}, $i.e.$, $\mathcal{F}_{i}^{In3}= \mathcal{C}_1[\mathcal{F}_{i}^{sam},\mathcal{F}_{i}^{asf}]\oplus\mathcal{F}_{i}^{In1}$.

To further obtain powerful representations, we design a \textbf{gated cross feed-forward network (GCFFN)}. Specifically, the generated feature $\mathcal{F}_{i}^{In3}$ is spatially encoded via a layer normalization ($\mathbb{LN}(\cdot)$) and a 1$\times$1 point-wise convolution ($\mathcal{C}_3$), followed by a 3$\times$3 depth-wise separable convolution ($\mathcal{DC}_3$) to capture spatial correlations among features. Meanwhile, we apply a non-linear function and a gated mechanism to enhance the non-linear relationships among features. We then use another 3$\times$3 depth-wise separable convolution ($\mathcal{DC}_3$) to model spatial correlations, resulting in the feature $\mathcal{F}_{i}^{g1}$, $i.e.$,
\begin{equation}
	\begin{split}
           &{\mathcal{F}_{i}^{g1}=\mathcal{DC}_3(\Theta(\mathcal{DC}_3\mathcal{C}_{1}\mathbb{LN}(\mathcal{F}_{i}^{In3}))\otimes\mathcal{DC}_3\mathcal{C}_{1}\mathbb{LN}(\mathcal{F}_{i}^{In3}))}, \\
	\end{split}
\end{equation}
where $\Theta(\cdot)$ represents the GeLU function. Furthermore, we introduce the designed ASF to {further} enhance global perception capability, and integrate the features $\mathcal{F}_{i}^{g1}$ and $\mathcal{F}_{i}^{g2}(\mathcal{F}_{i}^{g2}=\mathbb{ASF}(\mathcal{C}_{1}\mathcal{F}_{i}^{In3})$) to obtain the feature {$\mathcal{F}_{i}^{geb}$} {with global properties}, {formula as:} $\mathcal{F}_{i}^{geb}=\mathcal{C}_1[\mathcal{F}_{i}^{g1},\mathcal{F}_{i}^{g2}]\oplus\mathcal{F}_{i}^{In3}.$

In high-quality polyp segmentation, sharp boundaries are equally indispensable. Therefore, \textbf{we construct the local enhancement block (LEB),} which adopts two sets of small-kernel {lightweight} convolutions to extract local spatial details from input features $\mathcal{F} _{i}^{e}$ and $\mathcal{S} _{i+1}^{m}$, {aiming} to refine the {boundaries} of {polyps}. Technically, input features $\mathcal{F} _{i}^{e}$ and $\mathcal{S} _{i+1}^{m}$ {are} first processed by a sequence of convolutions, which includes {a} 1$\times$1 convolution, a 3$\times$3 depthwise separable convolution, and {another} 1$\times$1 convolution, to obtain a local feature $\mathcal{F} _{i}^{l1}$, $i.e.$, $\mathcal{F} _{i}^{l1}= \mathcal{C}_1{\mathcal{DC}_3}\mathcal{C}_1[\mathcal{F} _{i}^{e},\mathcal{S} _{i+1}^{m}]$. Meanwhile, the second set of convolutions, comprising a 1$\times$1 convolution, a depthwise separable 5$\times$5 convolution, and another 1$\times$1 convolution, is applied to acquire local feature $\mathcal{F} _{i}^{l2}$, that is, $\mathcal{F} _{i}^{l2}= \mathcal{C}_1{\mathcal{DC}_5}\mathcal{C}_1[\mathcal{F} _{i}^{e},\mathcal{S} _{i+1}^{m}]$. Furthermore, the two local features $\mathcal{F} _{i}^{l1}$ and $\mathcal{F} _{i}^{l2}$ are aggregated to {generate} the feature $\mathcal{F} _{i}^{leb}$ with abundant spatial details, that is, $\mathcal{F}_{i}^{leb}= \mathcal{F}_{i}^{l1} \oplus \mathcal{F}_{i}^{l2}$. 

Finally, we integrate the global feature $\mathcal{F}_{i}^{geb}$ and the local feature $\mathcal{F}_{i}^{leb}$ to generate the feature $\mathcal{S}_{i}^{m}$, which contains abundant global and local contextual information, as follows:
\begin{equation}
	\begin{split}
&\mathcal{S}_{i}^{m}=\mathcal{C}_1[\mathcal{F}_{i}^{geb},\mathcal{F}_{i}^{leb}]\oplus\mathcal{F}_{i}^{In1}, i= 2,3,4,5. 
	\end{split}
\end{equation}

\begin{figure}
   	   	\centering\includegraphics[width=0.48\textwidth,height=3.3cm]{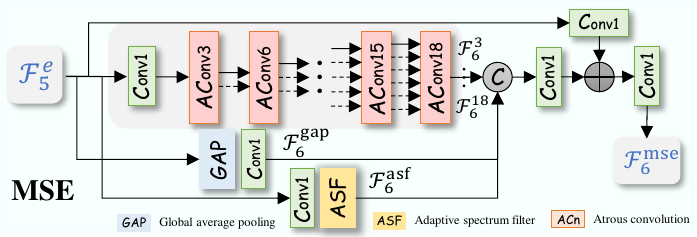}
   	   	\caption{Details of the multi-source semantic extractor.}
   	   	\label{mse_L}		
\end{figure}

\subsection{\textbf{Multi-source Semantic Extractor}}
\label{MFS}
Multi-source semantic extractor (MSE) is designed to facilitate coarse polyp localization in complex backgrounds by aggregating {rich} semantic cues from deep features. It uses an adaptive spectrum filter (ASF) and global average pooling (GAP) to obtain image-level semantics and produce higher-level features enriched with multi-source information. Technically, our MSE consists of $n$ ($n$=6) local branches and two global branches. For {the} local receptive field branches, we utilize atrous convolutions ($\mathcal{AC}_{3}^{k}$) with a 3$\times$3 kernel, where the dilation rate is {$k$ ($k=3n$)}. Atrous convolutions ($\mathcal{AC}_{3}^{k}$) effectively enlarge the receptive field without introducing extra parameters. We employ dense connections to enhance the correlation between semantic information from different local perspectives. Local receptive features $\{\mathcal{F}_{6}^{3n}\}_{n=1}^{6}$ are obtained using the following formula: {$\mathcal{F}_6^{3n}=\mathcal{AC}_{3}^{k}(\mathcal{C}_1\mathcal{F}_5+\sum_{n=2}^{6} \mathcal{F}_6^{3n-3})$}, where {$\mathcal{AC}_{3}^{k}$} represents the 3$\times$3 atrous convolution with a filling rate of {$k (k=3n)$}, $\mathcal{C}_{1}$ denotes the 1$\times$1 convolution. For the global receptive field branches, we first apply a global average pooling (GAP) operation to extract high-level semantic $\mathcal{F}_{6}^{gap}$ with global attributes, defined as $\mathcal{F}_{6}^{gap}=\mathcal{C}_1\mathbb{GAP} (\mathcal{F}_{5})$. Meanwhile, we adopt the adaptive spectrum filter (ASF) to capture the global spectrum information $\mathcal{F}_{6}^{asf}$, $i.e.$, $\mathcal{F}_{6}^{asf}=\mathbb{ASF} (\mathcal{F}_{5})$. Afterward, the obtained multi-source features are concatenated and processed with convolution operations to generate a coarse high-level semantic map $\mathcal{F}_{6}^{mse}$ with a single channel, which serves as guidance for the bottom features to identify the polyp location. The complete process is formulated as follows:
\begin{equation}
	\begin{split}
        &{\mathcal{F}_{6}^{mse}=\mathcal{C}_{1}(\mathcal{C}_{1}[\mathcal{F}_{6}^{3}, \mathcal{F}_{6}^{6}, ... , \mathcal{F}_{6}^{18},  \mathcal{F}_{6}^{gap}, \mathcal{F}_{6}^{asf}] \oplus \mathcal{C}_{1}\mathcal{F}_{5})}.\\
	\end{split}
\end{equation}

\begin{figure}
   	   	\centering\includegraphics[width=0.48\textwidth,height=4.2cm]{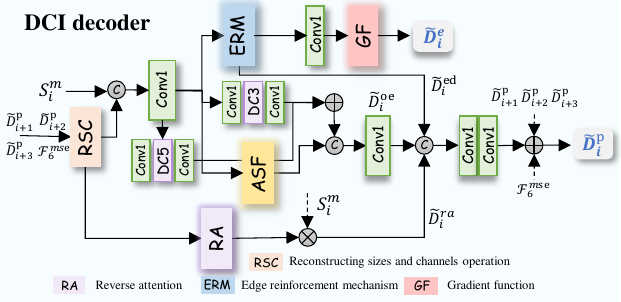}
   	   	\caption{Details of the dense cross-layer interaction decoder.}
   	   	\label{DCI_L}		
\end{figure}

\begin{table*}[t]
\renewcommand{\arraystretch}{1}
\setlength{\tabcolsep}{4pt}
	\centering
	\caption{Quantitative results on two polyp segmentation benchmarks. The best three results are shown in {\color{reda} \textbf{red}}, {\color{mygreen} \textbf{green}}, and {\color{myblue} \textbf{blue}}. ``$\uparrow$/$\downarrow$'' denotes that a higher/lower score is better. ``Ours-R'' or ``Ours-P'' present ResNet50 or PVTv2B3 as backbone.}
	\resizebox*{0.93\textwidth}{95mm}{
\begin{tabular}{c|cccccc|cccccc}
\hline \hline
\multirow{2}{*}{Methods} & \multicolumn{6}{c|}{CVC-300 (60 images)}                   & \multicolumn{6}{c}{CVC-ColonDB (380 images)}               \\
                         & \cellcolor{red!15} $Dic$$\uparrow$  & \cellcolor{red!15}$IoU$$\uparrow$  & \cellcolor{red!15}$F_{m}^{w}$$\uparrow$   & \cellcolor{red!15}$S_m$$\uparrow$    & \cellcolor{red!15}$E_m$$\uparrow$   & \cellcolor{red!15}$\mathcal{M}$$\downarrow$   & \cellcolor{red!15}$Dic$$\uparrow$  & \cellcolor{red!15}$IoU$$\uparrow$  & \cellcolor{red!15}$F_{m}^{w}$$\uparrow$   & \cellcolor{red!15}$S_m$$\uparrow$    & \cellcolor{red!15}$E_m$$\uparrow$   & \cellcolor{red!15}$\mathcal{M}$$\downarrow$   \\ \hline 
                         \multicolumn{13}{c}{\textbf{{CNN-based Polyp Segmentation Methods}}}                                                                                                                                                                                                             \\ \hline 
UNet \cite{UNet}                     & 0.710 & 0.627 & 0.684 & 0.843 & 0.847 & 0.022 & 0.504 & 0.436 & 0.491 & 0.710 & 0.692 & 0.059 \\ 
UNet++ \cite{UNet++}                  & 0.707 & 0.624 & 0.687 & 0.839 & 0.834 & 0.017 & 0.482 & 0.408 & 0.467 & 0.692 & 0.680 & 0.061 \\ 
SFA \cite{SFA}                     & 0.467 & 0.329 & 0.341 & 0.640 & 0.644 & 0.065 & 0.456 & 0.337 & 0.366 & 0.628 & 0.661 & 0.094 \\ 
ACSNet \cite{ACSNet}                  & 0.863 & 0.787 & 0.825 & 0.923 & 0.939 & 0.013 & 0.716 & 0.649 & 0.697 & 0.829 & 0.839 & \color{myblue}\textbf{0.039} \\ 
EUNet \cite{EUNet}                   & 0.837 & 0.765 & 0.805 & 0.904 & 0.919 & 0.015 & 0.704 & 0.631 & 0.684 & 0.821 & 0.840 & 0.052 \\ 
%SANet \cite{SANet}                   & 0.888 & 0.815 & 0.859 & 0.928 & \color{myblue}\textbf{0.962} & 0.008 & 0.753 & 0.670 & 0.726 & 0.837 & 0.869 & 0.043 \\ 
MSNet \cite{MSNet}         & 0.865 & 0.799 & 0.848 & 0.926 & 0.945 & 0.010 & 0.751 & 0.671 & 0.736 & 0.838 & \color{myblue}\textbf{0.872} & 0.041 \\ 
GLFRNet \cite{GLFRNet}         & 0.890 & 0.815 & 0.863 & 0.927 & 0.955 & \color{myblue}\textbf{0.007} & {0.756} & \color{myblue}\textbf{0.687} & \color{myblue}\textbf{0.745} & \color{myblue}\textbf{0.840} & 0.869 & \color{mygreen}\textbf{0.036} \\ 
DCRNet \cite{DCRNet}               & 0.856 & 0.788 & 0.830 & 0.921 & 0.955 & 0.010 & 0.704 & 0.631 & 0.684 & 0.821 & 0.840 & 0.052 \\ 
FTMFNet \cite{FTPS1}               & 0.890 & 0.811 & - & - & - & - & \color{reda}\textbf{0.796} & \color{reda}\textbf{0.705} & - & - & - & - \\ 
CFANet \cite{CFANet}           & \color{myblue}\textbf{0.893} &\color{myblue}\textbf{0.827} & \color{myblue}\textbf{0.875} & \color{myblue}\textbf{0.938} & \color{myblue}\textbf{0.962} & 0.008 & 0.743 & 0.665 & 0.728 & 0.835 & 0.869 & \color{myblue}\textbf{0.039} \\ 
%MEGANet \cite{MEGANet}         & 0.887 & 0.818 & 0.863 & 0.924 & 0.956 & 0.009 & \color{myblue}\textbf{0.756} & 0.681 & 0.730 & 0.831 & 0.863 & 0.045 \\ 
{UHANet} \cite{TCSVT4}         & \color{mygreen}\textbf{0.902} & \color{mygreen}\textbf{0.838} & \color{mygreen}\textbf{0.886} & \color{reda}\textbf{0.942} & \color{mygreen}\textbf{0.976} & \color{mygreen}\textbf{0.006} & \color{myblue}\textbf{0.769} & \color{mygreen}\textbf{0.695} & \color{mygreen}\textbf{0.755} & \color{mygreen}\textbf{0.849} & \color{mygreen}\textbf{0.892} & \color{mygreen}\textbf{0.036} \\ 
\rowcolor{cyan!8} \textbf{Ours-R}                   & \color{reda}\textbf{0.912} & \color{reda}\textbf{0.855} & \color{reda}\textbf{0.898} & \color{mygreen}\textbf{0.941} & \color{reda}\textbf{0.980} & \color{reda}\textbf{0.005} & \color{mygreen}\textbf{0.778} & \color{reda}\textbf{0.705} & \color{reda}\textbf{0.763} & \color{reda}\textbf{0.854} & \color{reda}\textbf{0.899} & \color{reda}\textbf{0.031} \\ \hline
\multicolumn{13}{c}{\textbf{{Transformer-based Polyp Segmentation Methods}}}                                                                                                                                                                                                             \\ \hline 
{SwinUNet \cite{Swin-Unet}}         & 0.879 & 0.805 & -  & - & - & - & 0.735 & 0.651 & - & - & - & - \\
PolypPVT \cite{PolypPVT}                & \color{myblue}\textbf{0.900} & 0.833 & \color{mygreen}\textbf{0.884} & \color{myblue}\textbf{0.935} & \color{reda}\textbf{0.973} & \color{mygreen}\textbf{0.007} & 0.735 & 0.666 & 0.724 & 0.834 & 0.859 & 0.038 \\
{TransFuse \cite{Transfuse}}         & \color{mygreen}\textbf{0.904} & \color{myblue}\textbf{0.838} & -  & - & - & - & 0.744 & 0.676 & - & - & - & - \\
{UTNet \cite{UTNet}}         & 0.806 & 0.733 & 0.778 & 0.882 & 0.924 & 0.016 & 0.676 & 0.600 & 0.656 & 0.799 & 0.855 & 0.041 \\
{TransUNet \cite{TransUNet}}         & 0.893 & 0.824 & -  & - & - & - & 0.781 & 0.699 & - & - & - & - \\
{Polyp-Mixer} \cite{TCSVT2}           & - & - & - & - & - & - & {0.791} & 0.706 & 0.768 & 0.862 & 0.893 & - \\ 
PPNet \cite{PPNet}        & 0.899 & \color{mygreen}\textbf{0.839} & {0.879} & \color{mygreen}\textbf{0.937} & \color{myblue}\textbf{0.966} & \color{reda}\textbf{0.006} & {0.791} & {0.726} & {0.776} & \color{myblue}\textbf{0.865} & {0.905} & \color{mygreen}\textbf{0.028} \\ 
{LSSNet} \cite{LSSNet}         & 0.884 & 0.815 & 0.852 & 0.930 & 0.943 & {0.009} & \color{mygreen}\textbf{0.816} & \color{myblue}\textbf{0.739} & \color{myblue}\textbf{0.791} & \color{mygreen}\textbf{0.875} & \color{myblue}\textbf{0.911} & {0.030} \\ 
{RUN} \cite{RUN}               & 0.879 & 0.818 & 0.845 & 0.932 & 0.943 & 0.010 & \color{myblue}\textbf{0.810} & \color{mygreen}\textbf{0.740} & {0.780} & \color{reda}\textbf{0.877} & 0.895 & \color{mygreen}\textbf{0.028} \\ 
{MSBP} \cite{MSBP}               & 0.897 & 0.828 & \color{myblue}\textbf{0.882} & 0.931 & 0.965 & \color{myblue}\textbf{0.008} & \color{myblue}\textbf{0.810} & {0.731} & \color{reda}\textbf{0.810} & {0.861} & \color{mygreen}\textbf{0.912} & \color{myblue}\textbf{0.029} \\ 

\rowcolor{cyan!8} \textbf{Ours-P}                      & \color{reda}\textbf{0.909} & \color{reda}\textbf{0.846} & \color{reda}\textbf{0.889} & \color{reda}\textbf{0.942} &\color{mygreen}\textbf{0.968} & \color{reda}\textbf{0.006} & \color{reda}\textbf{0.824} & \color{reda}\textbf{0.748} & \color{mygreen}\textbf{0.804} & \color{mygreen}\textbf{0.875} & \color{reda}\textbf{0.915} & \color{reda}\textbf{0.025} \\ \hline \hline
\end{tabular}}
\label{TWO results}
\end{table*}

\begin{table*}[]
\renewcommand{\arraystretch}{0.8}
\setlength{\tabcolsep}{2.5pt}
	\centering
	\caption{Quantitative results on three polyp segmentation benchmarks.}
	\resizebox*{0.93\textwidth}{75mm}{
\begin{tabular}{c|cccccc|cccccc|cccccc}
\hline \hline
\multirow{2}{*}{Methods} & \multicolumn{6}{c|}{ETIS-Larib (196 images)}        & \multicolumn{6}{c|}{Kvasir (100 images)}  & \multicolumn{6}{c}{CVC-ClinicDB (62 images)}                    \\
                         & \cellcolor{red!15} $Dic$$\uparrow$  & \cellcolor{red!15}$IoU$$\uparrow$  & \cellcolor{red!15}$F_{m}^{w}$$\uparrow$   & \cellcolor{red!15}$S_m$$\uparrow$    & \cellcolor{red!15}$E_m$$\uparrow$   & \cellcolor{red!15}$\mathcal{M}$$\downarrow$   & \cellcolor{red!15}$Dic$$\uparrow$  & \cellcolor{red!15}$IoU$$\uparrow$  & $F_{m}^{w}$$\uparrow$   & \cellcolor{red!15}$S_m$$\uparrow$    & \cellcolor{red!15}$E_m$$\uparrow$   & \cellcolor{red!15}$\mathcal{M}$$\downarrow$ & \cellcolor{red!15}$Dic$$\uparrow$  & \cellcolor{red!15}$IoU$$\uparrow$  & \cellcolor{red!15}$F_{m}^{w}$$\uparrow$   & \cellcolor{red!15}$S_m$$\uparrow$    & \cellcolor{red!15}$E_m$$\uparrow$   & \cellcolor{red!15}$\mathcal{M}$$\downarrow$   \\ \hline
\multicolumn{19}{c}{\textbf{{CNN-based Polyp Segmentation Methods}}}                                                                                                                                                                                                             \\ \hline 
UNet \cite{UNet}                    & 0.398 & 0.335 & 0.366 & 0.684 & 0.643 & 0.036 & 0.818 & 0.746 & 0.794 & 0.858 & 0.881 & 0.055 & 0.823 & 0.875 & 0.811 & 0.889 & 0.913 & 0.019 \\ 
UNet++ \cite{UNet++}                  & 0.401 & 0.344 & 0.390 & 0.683 & 0.629 & 0.035 & 0.821 & 0.743 & 0.808 & 0.862 & 0.886 & 0.048 & 0.794 & 0.728 & 0.785 & 0.873 & 0.891 & 0.022\\ 
SFA \cite{SFA}                     & 0.297 & 0.217 & 0.231 & 0.557 & 0.531 & 0.109 & 0.723 & 0.611 & 0.670 & 0.782 & 0.834 & 0.075 & 0.700 & 0.607 & 0.647 & 0.793 & 0.840 & 0.042 \\ 
ACSNet \cite{ACSNet}                  & 0.579 & 0.509 & 0.530 & 0.754 & 0.737 & 0.059 & 0.898 & 0.838 & 0.882 & 0.920 & 0.941 & 0.032  & 0.882	&0.826	&0.873	&0.927	&0.947	&0.011 \\
EUNet \cite{EUNet}                   & 0.687 & 0.609 & 0.636 & 0.793 & 0.807 & 0.067 & \color{myblue}\textbf{0.908} & 0.854 & 0.893 & 0.917 & 0.951 & \color{myblue}\textbf{0.028} & 0.902	& 0.846	& 0.891	& 0.936	& 0.959	& 0.011 \\ 
%SANet \cite{SANet}                   & \color{mygreen}\textbf{0.750} & 0.654 & 0.685 & 0.849 & 0.881 & \color{mygreen}\textbf{0.015} & 0.904 & 0.847 & 0.892 & 0.915 & 0.949 & 0.028 & 0.916	& 0.859	& 0.909	& 0.939	& 0.971	& 0.012 \\ 
MSNet \cite{MSNet}         & 0.723 & 0.652 & 0.677 & \color{mygreen}\textbf{0.845} & 0.876 & \color{myblue}\textbf{0.020} & 0.905 & 0.849 & 0.892 & \color{myblue}\textbf{0.923} & 0.947 & \color{myblue}\textbf{0.028} & 0.918	& \color{myblue}\textbf{0.869}	& 0.913	& 0.946	& \color{myblue}\textbf{0.973}	& \color{mygreen}\textbf{0.008} \\ 
GLFRNet \cite{GLFRNet}         & 0.709 & 0.634 & 0.660 & \color{myblue}\textbf{0.823} & 0.840 & 0.024 & 0.906 & 0.855 & 0.895 & 0.916 & 0.940 & 0.032  & \color{mygreen}\textbf{0.928}	& \color{reda}\textbf{0.883}	& \color{myblue}\textbf{0.918}	& \color{myblue}\textbf{0.949}	& 0.965	&\color{reda}\textbf{0.007} \\ 
DCRNet \cite{DCRNet}               & 0.556 & 0.496 & 0.506 & 0.736 & 0.773 & 0.096 & 0.886 & 0.825 & 0.868 & 0.911 & 0.935 & 0.035 & 0.896	& 0.844	& 0.890	& 0.933	& 0.968	& \color{myblue}\textbf{0.010}\\
FTMFNet \cite{FTPS1}               & \color{reda}\textbf{0.775} & \color{reda}\textbf{0.688} & - & - & - & - & 0.862 & 0.785 & - & - & - & - & 0.860	& 0.781	& -	& -	& -	& {-}\\
CFANet \cite{CFANet}           & {0.732} & {0.655} & \color{myblue}\textbf{0.693} & \color{mygreen}\textbf{0.845} & \color{myblue}\textbf{0.881} & \color{reda}\textbf{0.014} & \color{reda}\textbf{0.915} & \color{reda}\textbf{0.861} & \color{mygreen}\textbf{0.903} & \color{mygreen}\textbf{0.924} & \color{mygreen}\textbf{0.956} & \color{reda}\textbf{0.023} & \color{reda}\textbf{0.932}	& \color{reda}\textbf{0.883}	& \color{reda}\textbf{0.924}	& \color{mygreen}\textbf{0.950}	& \color{reda}\textbf{0.981}	& \color{reda}\textbf{0.007} \\ 
{UHANet} \cite{TCSVT4}               & \color{myblue}\textbf{0.746} & \color{myblue}\textbf{0.670} & \color{mygreen}\textbf{0.709} & \color{reda}\textbf{0.854} & \color{reda}\textbf{0.905} & 0.023 & \color{myblue}\textbf{0.908} & \color{myblue}\textbf{0.857} & \color{myblue}\textbf{0.901} & \color{reda}\textbf{0.926} & \color{myblue}\textbf{0.955} & \color{mygreen}\textbf{0.026} & \color{myblue}\textbf{0.927}	& \color{mygreen}\textbf{0.881}	& \color{mygreen}\textbf{0.923}	& \color{reda}\textbf{0.953}	& \color{mygreen}\textbf{0.979}	& \color{reda}\textbf{0.007}\\ 
\rowcolor{cyan!8} \textbf{Ours-R}                   & \color{mygreen}\textbf{0.747} & \color{mygreen}\textbf{0.674} & \color{reda}\textbf{0.712} & \color{mygreen}\textbf{0.845} & \color{mygreen}\textbf{0.884} & \color{mygreen}\textbf{0.015} & \color{mygreen}\textbf{0.914} & \color{mygreen}\textbf{0.859} & \color{reda}\textbf{0.905} & 0.920 & \color{reda}\textbf{0.959} & \color{reda}\textbf{0.023} & 0.910	&0.865	&0.906	&0.935 &0.972	&\color{myblue}\textbf{0.010} \\ \hline
\multicolumn{19}{c}{\textbf{{Transformer-based Polyp Segmentation Methods}}}                                                                                                                                                                                                  \\ \hline 
{SwinUNet \cite{Swin-Unet}}         & 0.680 & 0.595 & - & - & - & - & 0.887 & 0.828 & - & - & - & - & 0.892 & 0.837 & - & -	& -	& - \\ 
PolypPVT \cite{PolypPVT}                & \color{myblue}\textbf{0.787} & {0.706} & \color{myblue}\textbf{0.750} & \color{myblue}\textbf{0.871} & \color{mygreen}\textbf{0.906} & \color{mygreen}\textbf{0.013} & {0.917} & {0.864} & \color{mygreen}\textbf{0.911} & {0.925} & {0.956} & \color{mygreen}\textbf{0.023} & \color{mygreen}\textbf{0.937}	& \color{mygreen}\textbf{0.889}	&\color{mygreen}\textbf{0.936}	&\color{reda}\textbf{0.949}	&\color{mygreen}\textbf{0.985}	&\color{reda}\textbf{0.006}\\ 
{UTNet \cite{UTNet}}         & 0.556 & 0.489 & 0.522 & 0.749 & 0.772 & \color{myblue}\textbf{0.022} & 0.862 & 0.803 & 0.843 & 0.886 & 0.911 & 0.042 & 0.860	& 0.818	& 0.856	& 0.908	& 0.963	& {0.017} \\ 
{TransUNet \cite{TransUNet}}         & 0.731 & 0.660 & - & - & - & - & 0.913 & 0.857 & - & - & - & - & \color{myblue}\textbf{0.935} & \color{myblue}\textbf{0.887} & - & -	& -	& - \\ 
{TransFuse \cite{Transfuse}}         & 0.737 & 0.661 & - & - & - & - & {0.918} & \color{myblue}\textbf{0.868} & - & - & - & - & {0.934} &{0.886} & - & -	& -	& - \\ 

{Polyp-Mixer} \cite{TCSVT2}               & 0.759 & 0.676 & 0.711 & {0.863} & 0.875 & - & 0.916 & 0.864 & {0.908} & \color{reda}\textbf{0.932} & \color{myblue}\textbf{0.959} & - & 0.908	& 0.856	& 0.902	& 0.943	& 0.963	& - \\
PPNet \cite{PPNet}        & {0.784} & \color{myblue}\textbf{0.716} & 0.743 & \color{myblue}\textbf{0.871} & {0.885} & \color{mygreen}\textbf{0.013} & \color{mygreen}\textbf{0.920} & \color{reda}\textbf{0.874} & \color{mygreen}\textbf{0.911} & \color{mygreen}\textbf{0.927} & 0.949 & \color{myblue}\textbf{0.024} & 0.921	& 0.878	& 0.913	& \color{mygreen}\textbf{0.947}	& 0.969	& \color{myblue}\textbf{0.008}\\ 
%FreqFormer \cite{FTPS2}               & - & - & - & - & - & - & {0.918} & 0.867 & - & - & - & \color{mygreen}\textbf{0.023} & \color{reda}\textbf{0.937}	& \color{reda}\textbf{0.897}	& -	& -	& -	& \color{myblue}\textbf{0.009}\\ 

{LSSNet} \cite{LSSNet}               & 0.765 & 0.698 & {0.716} & 0.859 & 0.842 & 0.025 & 0.911 & 0.866 & 0.895 & \color{myblue}\textbf{0.926} & 0.949 & 0.028 & 0.920	& 0.875	& {0.914}	& \color{myblue}\textbf{0.944}	& {0.971}	& {0.010}\\ 
{{RUN} \cite{RUN}}         & 0.757 & 0.694 & 0.703 & 0.856 & 0.841 & 0.019 & 0.909 & \color{myblue}\textbf{0.868} & 0.893 & 0.922 & 0.941 & 0.028 & {0.934} & \color{reda}\textbf{0.893} & 0.925 & 0.932	& 0.943	& \color{myblue}\textbf{0.008} \\ 
{{MSBP} \cite{MSBP}}         & \color{mygreen}\textbf{0.795} & \color{mygreen}\textbf{0.718} & \color{reda}\textbf{0.770} & \color{mygreen}\textbf{0.873} & \color{myblue}\textbf{0.899} & \color{reda}\textbf{0.012} & \color{myblue}\textbf{0.919} & \color{myblue}\textbf{0.868} & \color{reda}\textbf{0.916} & \color{myblue}\textbf{0.926} & \color{mygreen}\textbf{0.961} & \color{reda}\textbf{0.022} & \color{reda}\textbf{0.940} & \color{reda}\textbf{0.893} & \color{reda}\textbf{0.941} & \color{reda}\textbf{0.949}	& \color{reda}\textbf{0.987}	& \color{mygreen}\textbf{0.007} \\ 
\rowcolor{cyan!10} \textbf{Ours-P}                      & \color{reda}\textbf{0.804} & \color{reda}\textbf{0.728} & \color{mygreen}\textbf{0.763} & \color{reda}\textbf{0.877} & \color{reda}\textbf{0.907} & \color{reda}\textbf{0.012} & \color{reda}\textbf{0.921} & \color{mygreen}\textbf{0.870} & \color{myblue}\textbf{0.910} & {0.925} & \color{reda}\textbf{0.962} & \color{mygreen}\textbf{0.023} & {0.932}	&\color{mygreen}\textbf{0.889}	& \color{myblue}\textbf{0.927}	& \color{myblue}\textbf{0.944}	&\color{myblue}\textbf{0.977}	&\color{myblue}\textbf{0.008} \\ \hline \hline
\end{tabular}}
\label{Three result}
\end{table*}

\subsection{\textbf{Dense Cross-layer Interaction Decoder}}
\label{PMSD}
For the optimized features $\mathcal{F}_{6}^{mse}$ and \{$\mathcal{S}_{i}^{m}\}_{i=2}^{5}$, we introduce a \textbf{dense cross-layer interaction (DCI) decoder} (as depicted in Fig. \ref{DCI_L}) to aggregate diverse types of information, thereby enhancing polyp segmentation. On the one hand, the DCI decoder integrates diverse information ($i.e.$, high-level semantics, and low-level details) from multiple feature layers; on the other hand, it further mines potentially significant information within channels. Specifically, in the fifth stage, we first aggregate the features $\mathcal{F}_{6}^{mse}$ and $\mathcal{S}_{5}^{m}$ to obtain a comprehensive feature $\mathcal{D}_{5}^{cf}$, defined as $\mathcal{D}_{5}^{cf} = \mathcal{C}_{1}[\mathcal{S}_{5}^{m},\Phi(\mathcal{F}_{6}^{mse})]$, where $\Phi(\cdot)$ denotes a transformation that reconstructs the feature to match the same dimension and spatial size as $\mathcal{S}_{5}^{m}$. Then, we perform two parallel operations ($i.e.$, edge reinforcement, and object enhancement). Specifically, in the edge reinforcement process, we first model the edge details based on the feature $\mathcal{D}_{5}^{cf}$ using lightweight convolutions, denoted as:
\begin{equation}
	\begin{split}
        &\tilde{\mathcal{D}}_{5}^{ed}=\mathcal{C}_{1}[\mathbb{MSA}(\mathcal{C}_{1}{\mathcal{DC}_{3}}\mathcal{C}_{1}\mathcal{D}_{5}^{cf}),\mathbb{MSA}(\mathcal{C}_{1}{\mathcal{DC}_{5}}\mathcal{C}_{1}\mathcal{D}_{5}^{cf})],\\
	\end{split}
\end{equation}
where $\mathbb{MSA}(\cdot)$ is the multi-scale spatial attention, see formula 2 for details. Afterwards, based on the edge feature $\mathcal{D}_{5}^{ed}$, we obtain an edge map through convolution operations and then enhance it using a gradient function ($\mathbb{GF}(\cdot)$) \cite{FEDER}, that is, $\tilde{\mathcal{D}}_{5}^{e}$ = $\mathbb{GF} (\mathcal{C}_{1}\tilde{\mathcal{D}}_{5}^{ed})$. In the object enhancement stage, we begin by enhancing both the local-global {clues} of $\mathcal{D}_{5}^{cf}$ using an ASF and a series of convolutions with small kernels, resulting in the optimized feature $\tilde{\mathcal{D}}_{5}^{oe}$. Furthermore, we utilize a reverse attention map generated from the high-level semantic feature to obtain a reverse feature {$\tilde{\mathcal{D}}_{5}^{ra}$}, which helps enhance the consistency of polyps in complex backgrounds, as follows:
\begin{equation}
	\begin{split}
        &\tilde{\mathcal{D}}_{5}^{oe}=\mathcal{C}_{1}[\mathbb{ASF}(\mathcal{D}_{5}^{cf}),\mathcal{C}_{1}{\mathcal{DC}_{3}}\mathcal{C}_{1}\mathcal{D}_{5}^{cf}\oplus\mathcal{C}_{1}{\mathcal{DC}_{5}}\mathcal{C}_{1}\mathcal{D}_{5}^{cf}],\\
        &\tilde{\mathcal{D}}_{5}^{ra}= \Phi (1-\sigma_{s}(\mathcal{F}_{6}^{mse})\oplus1)\otimes\mathcal{D}_{5}^{cf},\\
	\end{split}
\end{equation}
where $\Phi(\cdot)$ is used to reconstruct the same number of channels and spatial size. Subsequently, the obtained features ($i.e.$, $\tilde{\mathcal{D}}_{5}^{ed}$, $\tilde{\mathcal{D}}_{5}^{oe}$, and $\tilde{\mathcal{D}}_{5}^{ra}$) {are concatenated and passed through} two 1$\times$1 convolutions to reduce the dimensionality to a single channel. In addition, we aggregate the high-level feature $\mathcal{F}_{6}^{mse}$ to generate the final prediction {$\tilde{\mathcal{D}}_{5}^{p}$}, which is formulated as:
\begin{equation}
	\begin{split}
        &{\tilde{\mathcal{D}}_{5}^{p} =\mathcal{C}_{1}\mathcal{C}_{1}[\tilde{\mathcal{D}}_{5}^{ed},\tilde{\mathcal{D}}_{5}^{ra},\tilde{\mathcal{D}}_{5}^{oe}] \oplus \mathcal{F}_{6}^{mse}}.\\
	\end{split}
\end{equation}

Subsequently, from the fourth to second stages, we adopt dense connections to integrate high-level features ($i.e.$, $\mathcal{F}_{6}^{mse}$, $\tilde{\mathcal{D}}_{i+1}^{p}$, $\tilde{\mathcal{D}}_{i+2}^{p}$, and $\tilde{\mathcal{D}}_{i+3}^{p}$) into the optimized feature $\mathcal{S}_{i}^{m}$, and gradually decode them to generate a prediction map $\tilde{\mathcal{D}}_{i}^{p}$ and an edge map $\tilde{\mathcal{D}}_{i}^{e}$. The detailed formulation is given as:
\begin{equation}
	\begin{split}
&{(\tilde{\mathcal{D}}_{i}^{p},\tilde{\mathcal{D}}_{i}^{e})=\mathbb{DCI}(\mathcal{S}_{i}^{m}, \mathcal{F}_{6}^{mse}, \tilde{\mathcal{D}}_{i+1}^{p}, \tilde{\mathcal{D}}_{i+2}^{p}, \tilde{\mathcal{D}}_{i+3}^{p})},\\
	\end{split}
\end{equation}
where $\mathbb{DCI}(\cdot)$ denotes the dense cross-layer interaction decoder, and the indices $i+2$ and $i+3$ are constrained to be less than or equal to 5. Through optimization of the proposed DCI decoder, the feature $\mathcal{S}_{i}^{m}$ generated by the designed SNP module incorporates rich semantic and detailed information, which benefits colon polyp segmentation.

\subsection{\textbf{Loss Function}}
\label{LF}
We optimize the model parameters by supervising both prediction maps ($\{\tilde{\mathcal{D}}_{i}^{p}\}_{i=2}^{5}$, $\mathcal{F}_{6}^{mse}$) and edge features ($\{\tilde{\mathcal{D}}_{i}^{e}\}_{i=2}^{5}$). Specifically, we employ the weighted binary cross-entropy (BCE) loss and the weighted intersection-over-union (IoU) loss. Additionally, we use the Dice loss to improve accuracy in edge supervision. The loss function is formulated as follows:
\begin{equation}
		\begin{split}         
&{\mathcal{L}_{total}=\mathcal{L}_{bce}^{w}\oplus\mathcal{L}_{iou}^{w}\oplus\mathcal{L}_{dice}},
        %\frac{1}{2^{4}}(\mathcal{L}_{bce}^{w}(\sigma_s(\mathcal{F}_{6}^{mse}),GT_p)\oplus \mathcal{L}_{iou}^{w}(\sigma_s(\mathcal{F}_{6}^{mse}),GT_p)) \\ 
       %&\quad\quad\quad \oplus \sum_{i=2}^{5}\frac{1}{2^{i-2}}(\mathcal{L}_{bce}^{w}(\sigma_s(\tilde{\mathcal{D}}_{i}^{p}),GT_p)\oplus \mathcal{L}_{iou}^{w}(\sigma_s(\tilde{\mathcal{D}_{i}^{p}}),GT_p)) \\
        %&\quad\quad\quad \oplus \sum_{i=2}^{5}\frac{1}{2^{i-1}}(\mathcal{L}_{dice}(\sigma_s(\tilde{\mathcal{D}}_{i}^{e}),GT_e)), \\
		\end{split}  
\end{equation}
where $\mathcal{L}_{bce}^{w}(\cdot,\cdot)$ and $\mathcal{L}_{iou}^{w}(\cdot,\cdot)$ denote the weighted binary cross-entropy {function} and intersection over union function, $\mathcal{L}_{dice}(\cdot,\cdot)$ is the dice {function}. $GT_p$ and $GT_e$ represent the polyp ground truth and the edge ground truth, respectively. 

\begin{table*}[]
\renewcommand{\arraystretch}{0.8}
\setlength{\tabcolsep}{2.5pt}
	\centering
	\caption{Efficiency analysis of {the proposed} ASGNet model and existing polyp segmentation methods.}
	\resizebox*{0.97\textwidth}{12mm}{
\begin{tabular}{c|c|c|c|c|c|c|c|c|c|c|c|c|c|c|c|c}
\hline
            Methods   & \cellcolor{red!15} UNet   & \cellcolor{red!15} UNet++  & \cellcolor{red!15}ACSNet  & \cellcolor{red!15} EUNet & \cellcolor{red!15}MSNet & \cellcolor{red!15}DCRNet & \cellcolor{red!15}UTNet & \cellcolor{red!15}TransUNet & \cellcolor{red!15}TransFuse & \cellcolor{red!15}GLFRNet & \cellcolor{red!15}CFANet & \cellcolor{red!15}LSSNet & \cellcolor{red!15}RUN & \cellcolor{red!15}MSBP & \cellcolor{red!15}\textbf{Ours-R} & \cellcolor{red!15}\textbf{Ours-P} \\ \hline \hline
Parameters (M) $\downarrow$ & 31.04  & 36.63     & 29.45  & 31.36 & {27.69} & 28.73 & 34.00 & 93.20 & 143.39  & 29.80   & \color{reda}\textbf{25.24} & 35.94 & 65.17 & \color{mygreen}\textbf{25.52} & \color{myblue}\textbf{26.97} & 47.47   \\ 
FLOPs (G) $\downarrow$     & 103.52 & 261.20   & 21.61  & 27.50 & \color{myblue}\textbf{16.93} & {17.27} & 27.10  & 24.70 & 156.55 & {23.84}   & 55.36 & 17.51 & 61.83 & \color{reda}\textbf{12.86} & \color{mygreen}\textbf{16.65} & 22.70   \\ 
Inference (FPS) $\uparrow$     & $\sim$ 8 & $\sim$ 7   & $\sim$ 22  & - & $\sim$ \color{myblue}\textbf{70} & $\sim$ 53 & $\sim$ 7  & - & $\sim$ 45& {-}   & $\sim$ 24  & $\sim$ \color{reda}\textbf{85}  & $\sim$ {15} & {-}  & $\sim$ \color{mygreen}\textbf{72} & $\sim$ 48   \\ \hline
\end{tabular}}
\label{Efficiency}
\end{table*}
\begin{figure*}[t]
    
	\centering\includegraphics[width=0.91\textwidth,height=8.2cm]{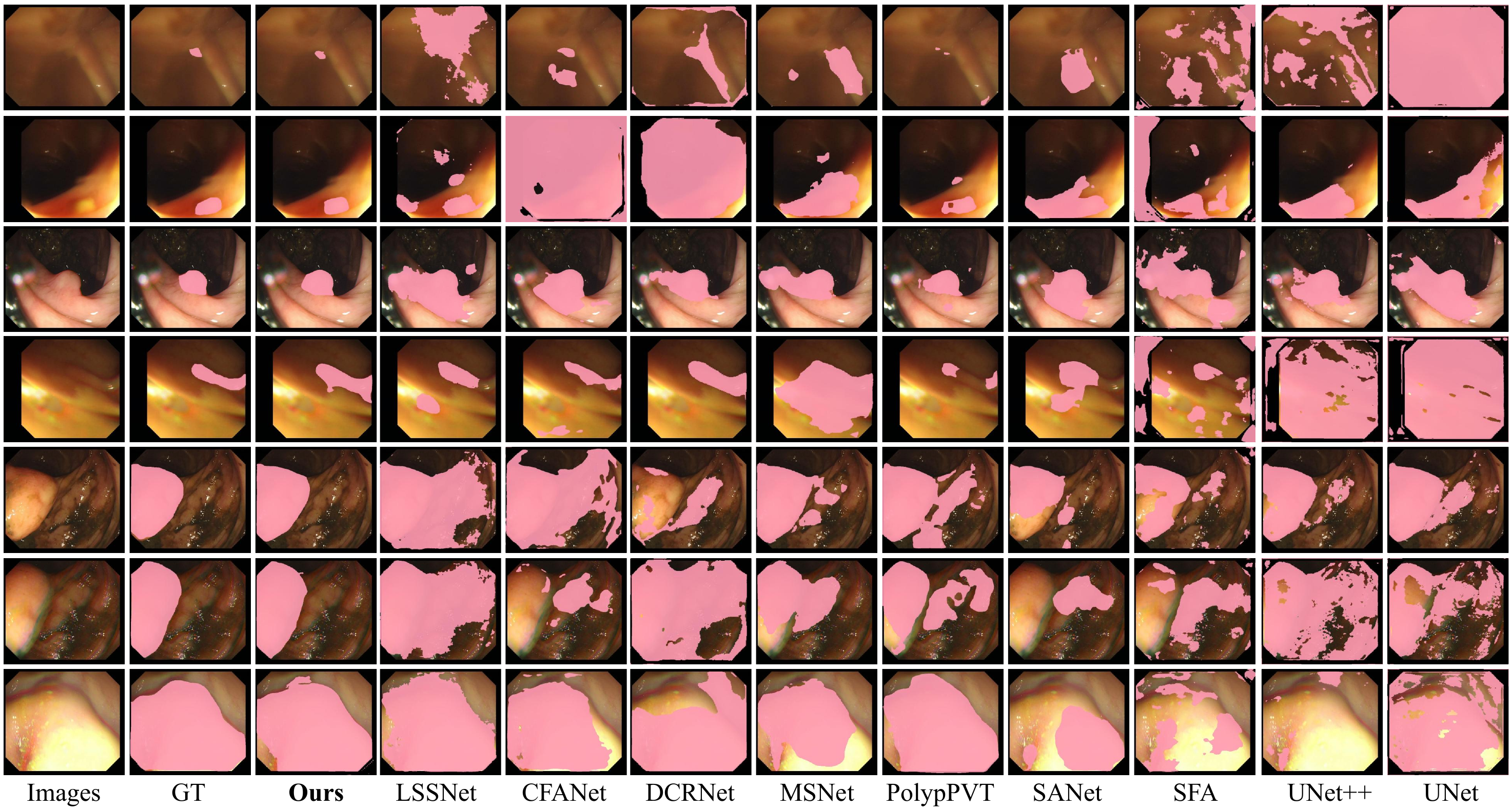}
    \captionsetup{font={small}, justification=raggedright}
	\caption{Qualitative results of the proposed ASGNet method and existing polyp segmentation {approaches}.}
	\label{visual_result}
\end{figure*}

\begin{table}[t]
\setlength{\tabcolsep}{1.5pt}
\centering
\caption{Ablation experiments on each component.}
\resizebox*{0.5\textwidth}{30mm}{
\begin{tabular}{c|cccc|cccc|cccc}
\hline \hline
\multirow{2}{*}{Num.} & \multicolumn{4}{c|}{{Component} Settings}                                             & \multicolumn{4}{c|}{CVC-300}   & \multicolumn{4}{c}{CVC-ColonDB} \\
                      & \multicolumn{1}{c}{Baseline} & \multicolumn{1}{c}{SNP} & \multicolumn{1}{c}{{MSE}} & DCI & \cellcolor{red!15} $Dic$$\uparrow$  & \cellcolor{red!15}$IoU$$\uparrow$  & \cellcolor{red!15} $F_{m}^{w}$$\uparrow$  & \cellcolor{red!15}$\mathcal{M}$$\downarrow$ &\cellcolor{red!15}$Dic$$\uparrow$  & \cellcolor{red!15} $IoU$$\uparrow$  & \cellcolor{red!15} $F_{m}^{w}$$\uparrow$  & \cellcolor{red!15}$\mathcal{M}$$\downarrow$  \\ \hline \hline
{(a)}                   & \multicolumn{1}{c|}{$\checkmark$}        & \multicolumn{1}{c|}{}  & \multicolumn{1}{c|}{}  &   & 0.852 & 0.766 & 0.807 & 0.010 & 0.729  & 0.642  & 0.700 & 0.039 \\ 
{(b)}                   & \multicolumn{1}{c|}{$\checkmark$}         & \multicolumn{1}{c|}{$\checkmark$} & \multicolumn{1}{c|}{}  &   & 0.877 & 0.803 & 0.844 & 0.008 & 0.764  & 0.681   & 0.736 & 0.033 \\ 
{(c)}                   & \multicolumn{1}{c|}{$\checkmark$}         & \multicolumn{1}{c|}{}  & \multicolumn{1}{c|}{$\checkmark$} &   & 0.873 & 0.791 & 0.831 & 0.008 & 0.752  & 0.667  & 0.720 & 0.034 \\ 
{(d)}                   & \multicolumn{1}{c|}{$\checkmark$}         & \multicolumn{1}{c|}{}  & \multicolumn{1}{c|}{}  & $\checkmark$ & 0.867 & 0.793 & 0.828 & 0.009 & 0.765  & 0.689  & 0.742 & 0.034 \\ 
{(e)}                   & \multicolumn{1}{c|}{$\checkmark$}         & \multicolumn{1}{c|}{$\checkmark$} & \multicolumn{1}{c|}{$\checkmark$} &   & 0.895 & 0.823 & 0.865 & 0.006 & 0.770  & 0.688  & 0.745 & 0.032 \\ 
{(f)}                   & \multicolumn{1}{c|}{$\checkmark$}         & \multicolumn{1}{c|}{$\checkmark$} & \multicolumn{1}{c|}{}  & $\checkmark$ & 0.895 & 0.827 & 0.870 & 0.007 & 0.763  & 0.689  & 0.743 & 0.033 \\ 
{(g)}                   & \multicolumn{1}{c|}{$\checkmark$}         & \multicolumn{1}{c|}{}  & \multicolumn{1}{c|}{$\checkmark$} & $\checkmark$ & 0.910 & 0.852 & 0.890 & 0.006 & \color{red}\textbf{0.778}  & 0.700  & 0.758 & 0.033 \\ 

\rowcolor{cyan!10} {(h)}                   & \multicolumn{1}{c|}{$\checkmark$}         & \multicolumn{1}{c|}{$\checkmark$} & \multicolumn{1}{c|}{$\checkmark$} & $\checkmark$ & \color{red}\textbf{0.912} & \color{red}\textbf{0.855} & \color{red}\textbf{0.898} & \color{red}\textbf{0.005} & \color{red}\textbf{0.778}  & \color{red}\textbf{0.705}  & \color{red}\textbf{0.763} & \color{red}\textbf{0.031} \\ \hline \hline
\end{tabular}}
\label{module study}
\end{table}

\begin{figure}[t]
	\centering\includegraphics[width=0.48\textwidth,height=3.7cm]{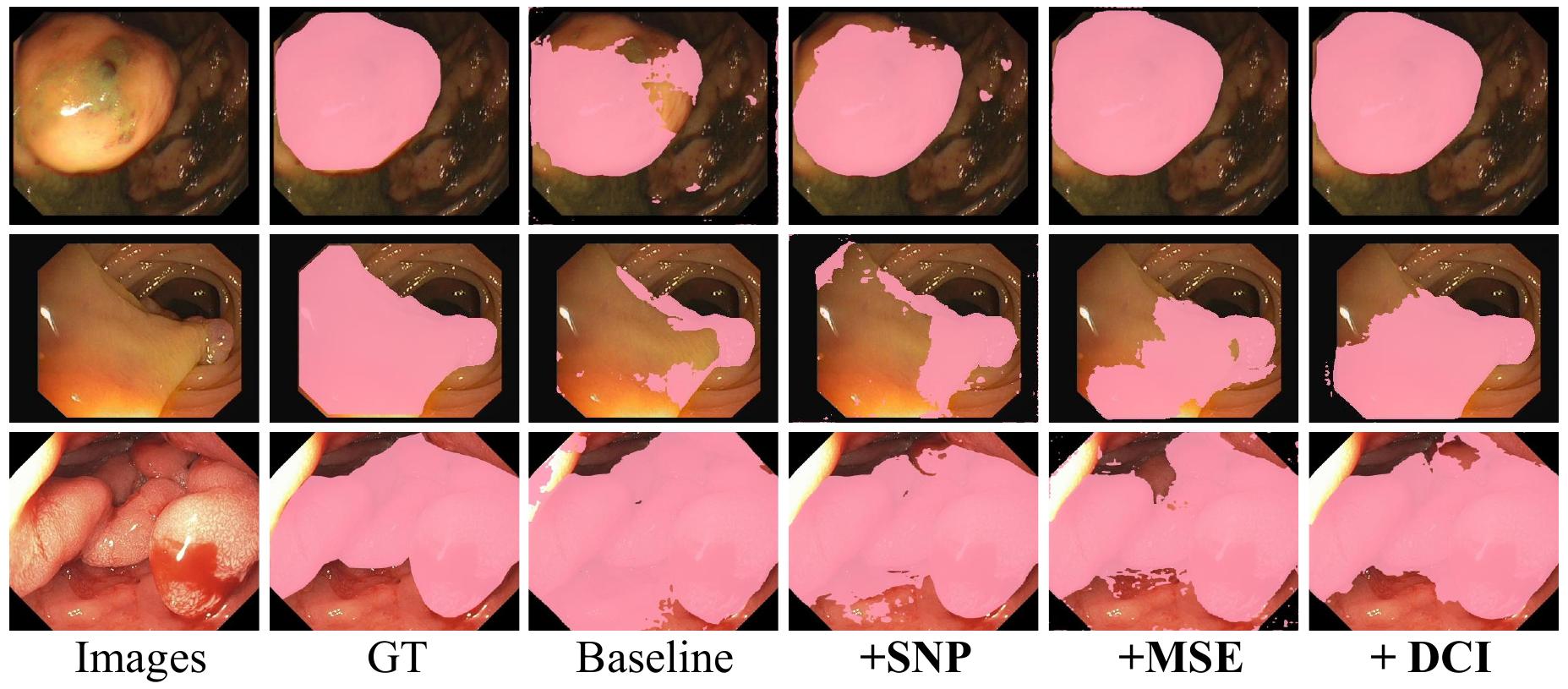}
	\captionsetup{font={small}, justification=raggedright}
	\caption{{Visual results of the effectiveness of each component}.}
        \label{all module}
\end{figure}

\section{Experiment}
\label{S4}
\subsection{\textbf{Datasets}}
To evaluate the effectiveness of our ASGNet, we utilize five widely-used polyp segmentation datasets, including CVC-300 \cite{CVC-300} (60 images), CVC-ColonDB \cite{ColonDB} (380 images), ETIS-Larib \cite{ETIS} (96 images), Kvasir \cite{kvasir} (100 images), and CVC-ClinicDB \cite{ClinicDB} (62 images). {For training, 550 images from the CVC-ClinicDB \cite{ClinicDB} dataset and 900 images from the Kvasir \cite{kvasir} dataset are used to learn the parameters of ASGNet}.

\subsection{\textbf{Experimental Details}}
We deploy the ASGNet model using the PyTorch framework and train it on four NVIDIA GTX 4090 GPUs. For the initial encoder, we use either a pre-trained ResNet50 \cite{ResNet} or PVTv2B3 \cite{PVTv2}. The Adam optimizer is employed with an initial learning rate of 1e-4, which decays by a factor of 0.1 every 80 epochs. We set the batch size to 40 and the input image size to 352 × 352. The training process is conducted for 120 epochs. Following \cite{CFANet,PolypPVT}, we apply data augmentation techniques such as horizontal flipping, and random cropping.

\subsection{\textbf{Evaluation Metrics}}
We utilize multiple standard metrics to evaluate the performance of our ASGNet method, including mean Dice similarity coefficient ($Dic$), mean Intersection over Union ($IoU$), weighted F-measure ($F_{m}^{w}$) \cite{Fm}, S-measure ($S_m$) \cite{Sm}, mean E-measure ($E_m$) \cite{Em}, mean absolute error ($\mathcal{M}$). Except for the $\mathcal{M}$ metric, higher values indicate better performance.

\begin{table}[t]
\setlength{\tabcolsep}{2pt}
	\centering
	\caption{ Structure analysis of the designed SNP module.}
	\resizebox*{0.48\textwidth}{23mm}{
\begin{tabular}{c|ccc|cccc|cccc}
\hline \hline
\multirow{3}{*}{Num.} & \multicolumn{3}{c|}{Structure Settings}                                   & \multicolumn{4}{c|}{\multirow{2}{*}{CVC-300}} & \multicolumn{4}{c}{\multirow{2}{*}{CVC-ColonDB}} \\ \cline{2-4}
                      & \multicolumn{1}{c|}{\multirow{2}{*}{LEB}} & \multicolumn{2}{c|}{GEB}      & \multicolumn{4}{c|}{}                                    & \multicolumn{4}{c}{}                                         \\  
                      & \multicolumn{1}{c|}{}                     & \multicolumn{1}{c}{SAM} & ASF & \cellcolor{red!15} $Dic$$\uparrow$  & \cellcolor{red!15}$IoU$$\uparrow$  & \cellcolor{red!15}$F_{m}^{w}$$\uparrow$  & \cellcolor{red!15}$\mathcal{M}$$\downarrow$ &\cellcolor{red!15}$Dic$$\uparrow$  & \cellcolor{red!15}$IoU$$\uparrow$  & \cellcolor{red!15} $F_{m}^{w}$$\uparrow$  & \cellcolor{red!15}$\mathcal{M}$$\downarrow$          \\ \hline \hline
{(a)}                   & \multicolumn{1}{c|}{$\checkmark$}                    & \multicolumn{1}{c|}{}   &     & 0.851        & 0.763        & 0.801        & 0.011       & 0.741         & 0.652         & 0.704         & 0.038        \\ 
{(b)}                   & \multicolumn{1}{c|}{$\checkmark$}                    & \multicolumn{1}{c|}{$\checkmark$}  &     & 0.864        & 0.779        & 0.818        & 0.010       & 0.744         & 0.665         & 0.714         & 0.037        \\ 
{(c)}                   & \multicolumn{1}{c|}{}                     & \multicolumn{1}{c|}{$\checkmark$}  & $\checkmark$   & 0.866        & 0.785        & 0.830        & 0.008       & 0.749         & 0.660         & 0.725         & 0.035        \\ 
\rowcolor{cyan!10} {(d)}                   & \multicolumn{1}{c|}{$\checkmark$}                    & \multicolumn{1}{c|}{$\checkmark$}  & $\checkmark$   & \color{red}\textbf{0.877}        & \color{red}\textbf{0.803}        & \color{red}\textbf{0.844}        & \color{red}\textbf{0.008}       & \color{red}\textbf{0.764}         & \color{red}\textbf{0.681}         & \color{red}\textbf{0.736}         & \color{red}\textbf{0.033}        \\ \hline \hline
\end{tabular}}
\label{module1_ag_c}
\end{table}

\begin{figure}[t]
    \centering
    \captionsetup{font={small}, justification=raggedright}

    \begin{minipage}[t]{0.48\textwidth}
        \centering
        \includegraphics[width=\textwidth,height=3.2cm]{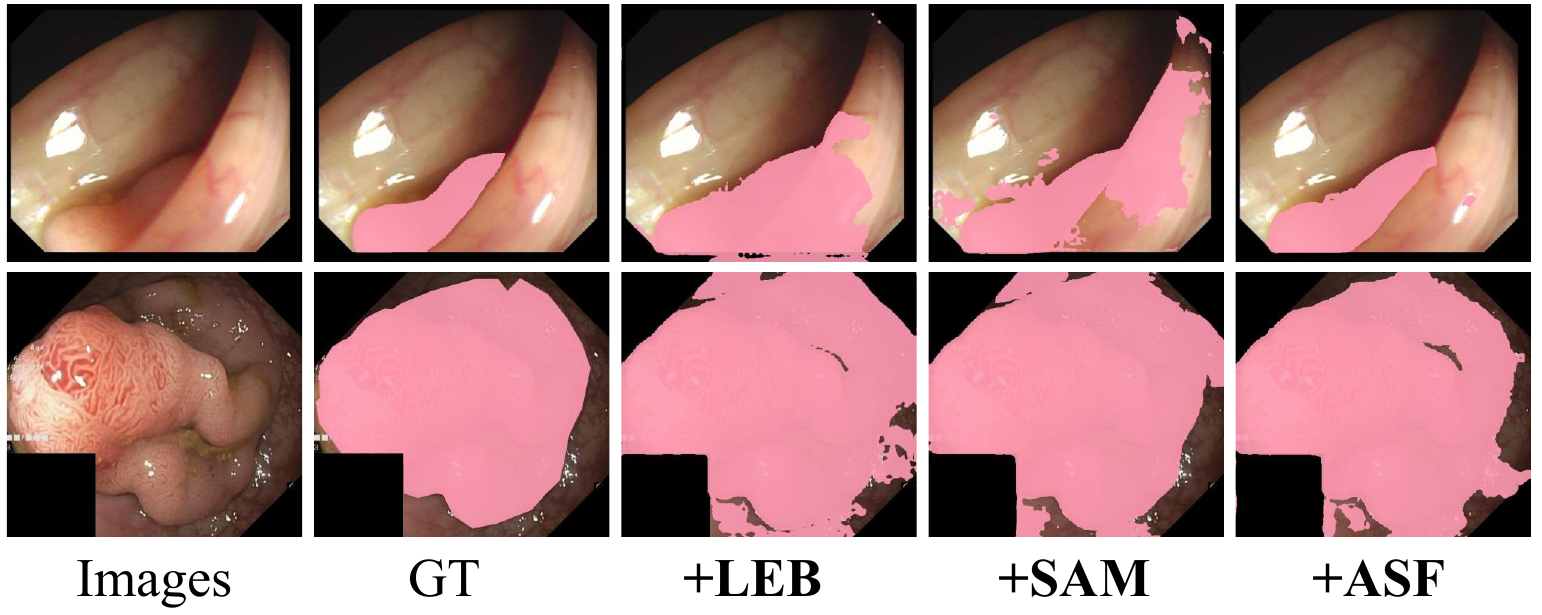}
        \caption{Visual results of ``+LEB'', ``+SAM'', and ``+ASF''.}
        \label{fig:visual_module1_a}
    \end{minipage}
    \hfill
    \
    \begin{minipage}[t]{0.48\textwidth}
        \centering
        \includegraphics[width=\textwidth,height=2.8cm]{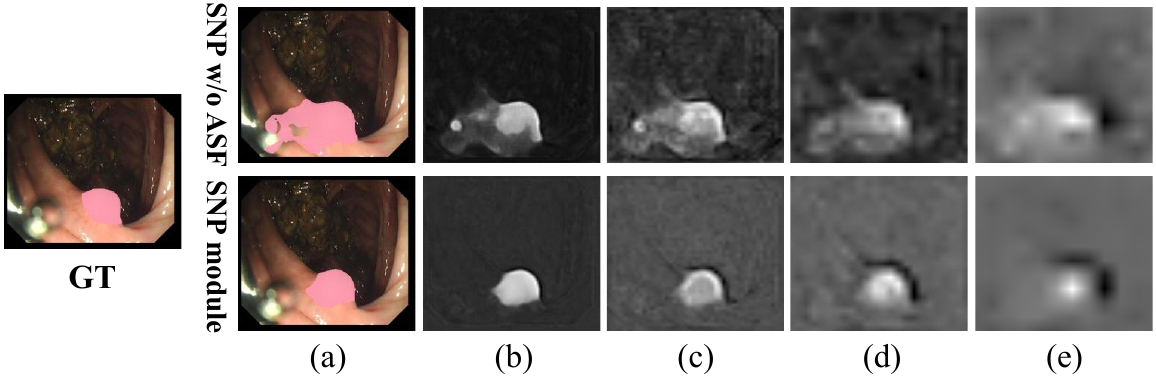}
        \caption{Visual results of our SNP. From left to right, (a) prediction maps, (b)-(e) visualization feature maps for stages 2 to 5.}
        \label{fig:visual_module1_b}
    \end{minipage}
\end{figure}

\begin{table}[t]
\setlength{\tabcolsep}{2pt}
	\centering
	\caption{ Comparison results of our SNP module compared to ``{Swin Transformer block (STB)'', ``Restormer block (RB)''}.}
	\resizebox*{0.48\textwidth}{18mm}{
\begin{tabular}{c|c|cccc|cccc}
\hline \hline
\multirow{2}{*}{Num.} & \multirow{2}{*}{Method} & \multicolumn{4}{c|}{CVC-300}   & \multicolumn{4}{c}{CVC-ColonDB} \\
                      &                         & \cellcolor{red!15} $Dic$$\uparrow$  & \cellcolor{red!15}$IoU$$\uparrow$  & \cellcolor{red!15}$F_{m}^{w}$$\uparrow$  & \cellcolor{red!15}$\mathcal{M}$$\downarrow$ &\cellcolor{red!15}$Dic$$\uparrow$  & \cellcolor{red!15}$IoU$$\uparrow$  & \cellcolor{red!15}$F_{m}^{w}$$\uparrow$  & \cellcolor{red!15}$\mathcal{M}$$\downarrow$   \\ \hline \hline
{(a)}                     & Base.+ STB \cite{Swin}             & \color{red}\textbf{0.879} & \color{red}\textbf{0.813} & \color{red}\textbf{0.861} & \color{red}\textbf{0.008} & 0.742  & 0.659  & 0.726 & 0.039 \\ 
{(b)}                     & Base.+ RB \cite{Restormer}             & 0.837 & 0.747 & 0.788 & 0.014 & 0.743  & 0.660   & 0.712 & 0.039 \\ 
\rowcolor{cyan!10} {(c)}                     & Base.+ SNP module         & 0.877 & 0.803 & 0.844 & \color{red}\textbf{0.008} & \color{red}\textbf{0.764}  & \color{red}\textbf{0.681}   & \color{red}\textbf{0.736} & \color{red}\textbf{0.033} \\ \hline \hline
\end{tabular}}
\label{module1_com_c}
\end{table}
\begin{figure}[t]
	\centering\includegraphics[width=0.48\textwidth,height=3.3cm]{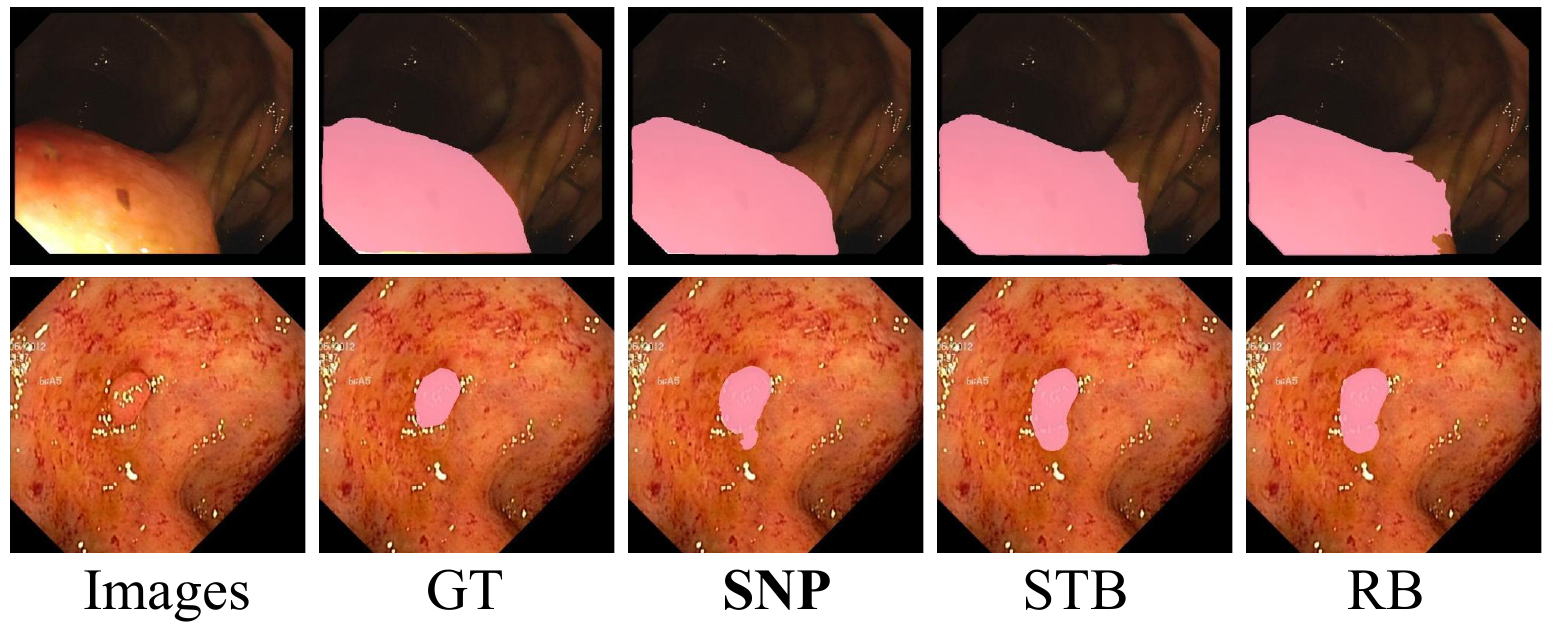}
	\captionsetup{font={small}, justification=raggedright}
	\caption{{Visual results of ``SNP'', ``STB'', and ``RB''}. }
        \label{visual module1-duibi}
\end{figure}

\subsection{\textbf{Comparison with State-of-the-Art Methods}}
We compare {our} ASGNet method with {21} SOTA polyp segmentation models, including UNet \cite{UNet}, UNet++ \cite{UNet++}, SFA \cite{SFA}, EUNet \cite{EUNet}, SANet \cite{SANet}, MSNet \cite{MSNet}, ACSNet \cite{ACSNet}, FTMFNet \cite{FTPS1}, PolypPVT \cite{PolypPVT}, GLFRNet \cite{GLFRNet}, DCRNet \cite{DCRNet}, CFANet \cite{CFANet}, UHANet \cite{TCSVT4}, SwinUNet \cite{Swin-Unet}, UTNet \cite{UTNet}, TransUNet \cite{TransUNet}, TransFuse \cite{Transfuse}, Polyp-Mixer \cite{TCSVT2}, PPNet \cite{PPNet}, LSSNet \cite{LSSNet}, {RUN \cite{RUN}, and MSBP \cite{MSBP}}. For a fair comparison, all {results} are either provided by the original authors or generated using publicly available source code. 

\subsubsection{\textbf{Quantitative Comparison}}
Tables \ref{TWO results} and \ref{Three result} summarize the quantitative results of our ASGNet and 21 existing methods on five polyp segmentation benchmarks, evaluated using six metrics ($i.e.$, $Dic$, $IoU$, $F_{m}^{w}$, $S_m$, $E_m$, and $\mathcal{M}$). To ensure a fair comparison, we categorize the methods based on CNN and Transformer backbones in Tables I and II. As shown in Table \ref{TWO results}, the proposed ASGNet model with different backbones ($i.e.$, ResNet-50 \cite{ResNet} or PVTv2B3 \cite{PVTv2}) achieves superior performance on both CVC-300 \cite{CVC-300} and CVC-ColonDB \cite{ColonDB} datasets. In particular, on CVC-300 \cite{CVC-300}, the predicted results of the ASGNet compared to the recent UHANet \cite{TCSVT4}, that is, 0.912 $vs.$ 0.902 in terms of $Dic$, 0.855 $vs.$ 0.838 in terms of $IoU$, 0.898 $vs.$ 0.886 in terms of $F_{m}^{w}$, 0.980 $vs.$ 0.976 in terms of $E_m$, and 0.005 $vs.$ 0.006 in terms of $\mathcal{M}$. Moreover, compared to the Transformer-based PolypPVT \cite{PolypPVT} and PPNet \cite{PPNet} approaches, the proposed ASGNet method achieves significant improve of 12.10\%, 12.31\%, 11.05\%, 4.92\%, 6.52\%, and 52\% over PolypPVT \cite{PolypPVT} and 4.17\%, 3.03\%, 3.61\%, 1.17\%, 1.11\%, and 12\% over PPNet \cite{PPNet} under the six evaluation metrics ($i.e.$, $Dic$, $IoU$, $F_{m}^{w}$, $S_m$, $E_m$, and $\mathcal{M}$) on CVC-ColonDB \cite{ColonDB}. Similarly, the performance advantage of our ASGNet method is demonstrated in ETIS-Larib \cite{ETIS} and Kvasir \cite{kvasir} datasets, as shown in Table \ref{Three result}. Furthermore, Table \ref{Efficiency} gives the model parameters, floating-point operations (FLOPs), and inference speed (FPS) of various polyp segmentation methods. As shown in Table \ref{Efficiency}, our ASGNet is competitive in terms of both the number of parameters, FLOPs, and FPS. In conclusion, the quantitative results provide strong evidence that the ASGNet model achieves superior accuracy and efficiency.

\begin{table}[t]
\centering
\begin{minipage}{0.48\textwidth}
\setlength{\tabcolsep}{2pt}
\centering
\caption{Comparison results of the proposed MSE compared to ``MSE w/o ASF.'', ``ASPP'', and ``RFB''.}
\resizebox{\textwidth}{12mm}{
\begin{tabular}{c|c|cccc|cccc}
\hline \hline
\multirow{2}{*}{Num.} & \multirow{2}{*}{Method} & \multicolumn{4}{c|}{CVC-300} & \multicolumn{4}{c}{CVC-ColonDB} \\
 & & \cellcolor{red!15} $Dic$$\uparrow$  & \cellcolor{red!15}$IoU$$\uparrow$  & \cellcolor{red!15}$F_{m}^{w}$$\uparrow$  & \cellcolor{red!15}$\mathcal{M}$$\downarrow$ &\cellcolor{red!15}$Dic$$\uparrow$  & \cellcolor{red!15}$IoU$$\uparrow$  & \cellcolor{red!15}$F_{m}^{w}$$\uparrow$  & \cellcolor{red!15}$\mathcal{M}$$\downarrow$ \\ \hline \hline
(a) & Base.+ {MSE} w/o ASF & 0.849 & 0.753 & 0.796 & 0.010 & 0.733 & 0.638 & 0.692 & 0.037 \\
(b) & Base.+ ASPP \cite{ASPP} & 0.867 & 0.785 & \color{red}\textbf{0.832} & \color{red}\textbf{0.008} & 0.747 & 0.653 & 0.716 & 0.036 \\
(c) & Base.+ RFB \cite{RFB} & 0.870 & 0.789 & \color{red}\textbf{0.832} & \color{red}\textbf{0.008} & 0.743 & 0.649 & 0.706 & 0.037 \\
\rowcolor{cyan!10}(d) & Base.+ MSE & \color{red}\textbf{0.873} & \color{red}\textbf{0.791} & 0.831 & \color{red}\textbf{0.008} & \color{red}\textbf{0.751} & \color{red}\textbf{0.667} & \color{red}\textbf{0.720} & \color{red}\textbf{0.034} \\ \hline \hline
\end{tabular}}
\label{module2}
\end{minipage}
\hfill
\begin{minipage}{0.48\textwidth}
\setlength{\tabcolsep}{2pt}
\centering
\caption{Filling rates analysis of our MSE.}
\resizebox{\textwidth}{12mm}{
\begin{tabular}{c|c|cccc|cccc}
\hline \hline
\multirow{2}{*}{Num.} & \multirow{2}{*}{Filling Rates} & \multicolumn{4}{c|}{CVC-300} & \multicolumn{4}{c}{CVC-ColonDB} \\
 & & \cellcolor{red!15} $Dic$$\uparrow$  & \cellcolor{red!15}$IoU$$\uparrow$  & \cellcolor{red!15}$F_{m}^{w}$$\uparrow$  & \cellcolor{red!15}$\mathcal{M}$$\downarrow$ &\cellcolor{red!15}$Dic$$\uparrow$  & \cellcolor{red!15}$IoU$$\uparrow$  & \cellcolor{red!15}$F_{m}^{w}$$\uparrow$  & \cellcolor{red!15}$\mathcal{M}$$\downarrow$ \\ \hline \hline
(a) & (1,1,1,1,1,1) & 0.857 & 0.775 & 0.815 & 0.011 & 0.750 & 0.665 & 0.716 & 0.039 \\
(b) & (1,2,3,4,5,6) & 0.872 & 0.779 & 0.830 & 0.009 & 0.745 & 0.661 & 0.709 & 0.036 \\
(c) & (2,4,6,8,12,16) & 0.848 & 0.763 & 0.798 & 0.012 & \color{red}\textbf{0.754} & 0.661 & 0.710 & 0.037 \\
\rowcolor{cyan!10}(d) & (3,6,9,12,15,18) & \color{red}\textbf{0.873} & \color{red}\textbf{0.791} & \color{red}\textbf{0.831} & \color{red}\textbf{0.008} & 0.751 & \color{red}\textbf{0.667} & \color{red}\textbf{0.720} & \color{red}\textbf{0.034} \\ \hline \hline
\end{tabular}}
\label{module2-fr}
\end{minipage}
\end{table}

\begin{figure}[t]
	\centering\includegraphics[width=0.48\textwidth,height=2.8cm]{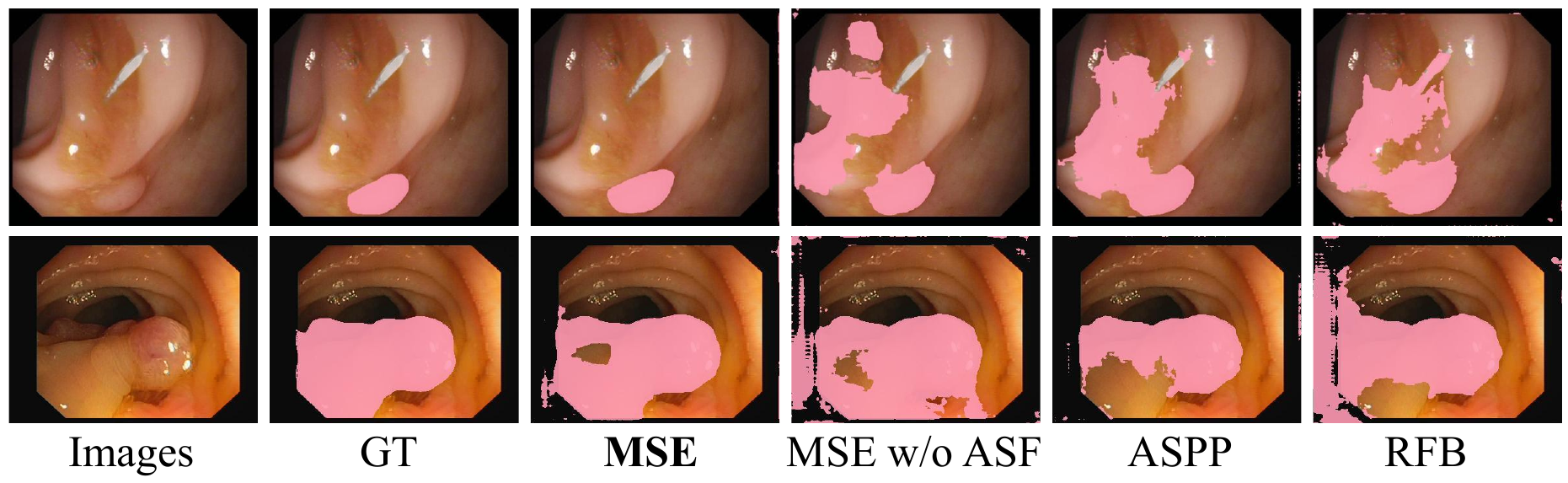}
	\captionsetup{font={small}, justification=raggedright}
	\caption{{Visual results of ``MSE'', ``MSE w/o ASF'', ``Atrous spatial pyramid pooling (ASPP)'' and ``Receptive field block (RFB)''}. }
        \label{visual module2-duibi}
\end{figure}

\begin{table}[t]
\setlength{\tabcolsep}{2pt}
	\centering
	\caption{Ablation analysis of the proposed DCI decoder.}
	\resizebox*{0.48\textwidth}{20mm}{
\begin{tabular}{c|c|cccc|cccc}
\hline \hline
\multirow{2}{*}{Num.} & \multirow{2}{*}{Method} & \multicolumn{4}{c|}{CVC-300}   & \multicolumn{4}{c}{CVC-ColonDB} \\
                      &                         & \cellcolor{red!15} $Dic$$\uparrow$  & \cellcolor{red!15}$IoU$$\uparrow$  & \cellcolor{red!15}$F_{m}^{w}$$\uparrow$  & \cellcolor{red!15}$\mathcal{M}$$\downarrow$ &\cellcolor{red!15}$Dic$$\uparrow$  & \cellcolor{red!15}$IoU$$\uparrow$  & \cellcolor{red!15}$F_{m}^{w}$$\uparrow$  & \cellcolor{red!15}$\mathcal{M}$$\downarrow$   \\ \hline \hline
{(a)}                     & Base.+ DCI w/o ASF    & 0.849 & 0.753 & 0.796 & 0.010 & 0.733  & 0.638  & 0.692 & 0.037 \\ 
{(b)}                     & Base.+ DCI w/o Edge    & 0.837 & 0.761 & 0.786 & 0.013 & 0.736  & 0.653  & 0.688 & 0.039 \\ 
\rowcolor{cyan!10}{(c)}                     & Base.+ DCI          & \color{red}\textbf{0.867} & \color{red}\textbf{0.793} & \color{red}\textbf{0.828} & \color{red}\textbf{0.009} & \color{red}\textbf{0.765}  & \color{red}\textbf{0.690}  & \color{red}\textbf{0.737} & \color{red}\textbf{0.034} \\ \hline \hline
\end{tabular}}
\label{module3}
\end{table}
\begin{figure}[t]
	\centering\includegraphics[width=0.48\textwidth,height=3.3cm]{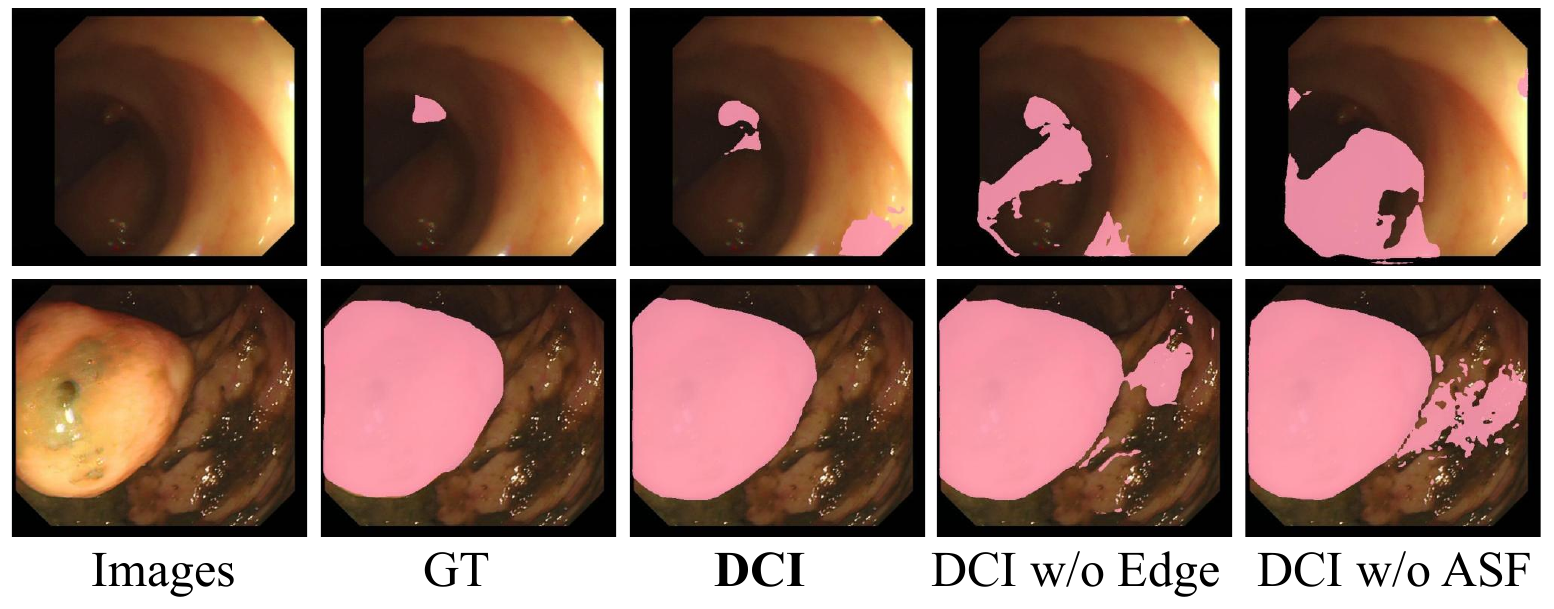}
	\captionsetup{font={small}, justification=raggedright}
	\caption{ Visual results of the DCI internal structure. }
	\label{visual module3}
        %\vspace{-1.5em}
\end{figure}

\begin{table}[t]
\setlength{\tabcolsep}{1.5pt}
	\centering
	\caption{Hyper-parameter analysis of channels.}
	\resizebox*{0.48\textwidth}{23mm}{
\begin{tabular}{c|c|c|c|cccc|cccc}
\hline \hline
\multirow{2}{*}{Num.} & Settings         & \multirow{2}{*}{\begin{tabular}[c]{@{}c@{}}Parameters\\ (M)\end{tabular}} & \multirow{2}{*}{\begin{tabular}[c]{@{}c@{}}FLOPs\\ (G)\end{tabular}} & \multicolumn{4}{c|}{CVC-300}  & \multicolumn{4}{c}{CVC-ColonDB} \\ \cline{2-2} 
                      & Decoder channels &                                                                           &                                                                      & \cellcolor{red!15} $Dic$$\uparrow$  & \cellcolor{red!15}$IoU$$\uparrow$  & \cellcolor{red!15}$F_{m}^{w}$$\uparrow$  & \cellcolor{red!15}$\mathcal{M}$$\downarrow$ &\cellcolor{red!15}$Dic$$\uparrow$  & \cellcolor{red!15}$IoU$$\uparrow$  & \cellcolor{red!15}$F_{m}^{w}$$\uparrow$  & \cellcolor{red!15}$\mathcal{M}$$\downarrow$   \\ \hline \hline
{(a)}                     & 32               & \color{red}\textbf{24.13}                                                                     & \color{red}\textbf{11.07}                                                                & 0.883 & 0.818 & 0.859 & 0.007 & \color{red}\textbf{0.785}  & \color{red}\textbf{0.710}   & 0.762 & \color{red}\textbf{0.031} \\ 
{(b)}                     & 64               & 25.28                                                                     & 13.22                                                                & 0.891 & 0.827  & 0.866 & 0.006 & 0.761  & 0.684  & 0.740 & 0.034 \\ 
{(c)}                     & 128              & 29.18                                                                     & 21.36                                                               & 0.904 & 0.842 & 0.889 & 0.005 & 0.782  & 0.706  & \color{red}\textbf{0.766} & \color{red}\textbf{0.031} \\ 
{(d)}                     & 160              & 31.94                                                                    & 27.36                                                               & 0.896 & 0.833 & 0.874 & 0.006 & 0.775  & 0.703  & 0.760 & \color{red}\textbf{0.031} \\ 
\rowcolor{cyan!10}{(e)}                     & 96              & 26.97                                                                     & 16.65                                                                & \color{red}\textbf{0.913} & \color{red}\textbf{0.853} & \color{red}\textbf{0.895} & \color{red}\textbf{0.005} & 0.778  & 0.705  & 0.763 & \color{red}\textbf{0.031} \\ \hline \hline
\end{tabular}}
\label{channels}
\end{table}

\subsubsection{\textbf{Qualitative Comparison}}
Fig. \ref{visual_result} presents the visual results of our ASGNet and nine existing segmentation models on complex intestinal scenes, where polyp objects are arranged in ascending order of scale. It can be observed that our ASGNet model accurately localizes polyps across various scales ($i.e.$, small, medium, {and} large objects), producing prediction results with sharp boundaries. In contrast, recent methods ($e.g.$, LSSNet \cite{LSSNet}, CFANet \cite{CFANet}, DCRNet \cite{DCRNet}, and PolypPVT \cite{PolypPVT}) tend to produce prediction error or incomplete segmentation when confronted with challenging scenes. This is because spectral information from the frequency domain, with its global perception, enhances pixel features from the spatial domain  and enables more accurate polyp inference.
\begin{table}[t]
\setlength{\tabcolsep}{3.5pt}
	\centering
	\caption{Hyper-parameter analysis of input sizes.}
	\resizebox*{0.48\textwidth}{24mm}{
\begin{tabular}{c|c|c|cccc|cccc}
\hline \hline
\multirow{2}{*}{Num.} & Settings   & \multirow{2}{*}{\begin{tabular}[c]{@{}c@{}}FLOPs\\ (G)\end{tabular}} & \multicolumn{4}{c|}{CVC-300}  & \multicolumn{4}{c}{CVC-ColonDB} \\ \cline{2-2} \cline{4-11} 
                      & input sizes &                                                                      & \cellcolor{red!15} $Dic$$\uparrow$  & \cellcolor{red!15}$IoU$$\uparrow$  & \cellcolor{red!15}$F_{m}^{w}$$\uparrow$  & \cellcolor{red!15}$\mathcal{M}$$\downarrow$ &\cellcolor{red!15}$Dic$$\uparrow$  & \cellcolor{red!15}$IoU$$\uparrow$  & \cellcolor{red!15}$F_{m}^{w}$$\uparrow$  & \cellcolor{red!15}$\mathcal{M}$$\downarrow$   \\ \hline \hline
{(a)}                     & 256$\times$256    & \color{red}\textbf{8.81}                                                                & 0.878 & 0.815 & 0.855 & 0.008 & \color{red}\textbf{0.787}  & \color{red}\textbf{0.708}  & \color{red}\textbf{0.764} & \color{red}\textbf{0.031} \\ 
{(b)}                     & 288$\times$288    & 11.15                                                                & 0.906 & 0.841  & 0.884 & 0.006 & 0.780  & 0.705  & 0.760 & \color{red}\textbf{0.031} \\ 
{(c)}                     & 384$\times$384    & 19.82                                                                & 0.898 & 0.835 & 0.876 & 0.007 & 0.781  & 0.707  & 0.758 & 0.032 \\ 
{(d)}                     & 416$\times$416    & 23.26                                                                & 0.899 & 0.835 & 0.878 & 0.007 & 0.780  & 0.705  & 0.763 & \color{red}\textbf{0.031} \\ 
\rowcolor{cyan!10}{(e)}                     & 352$\times$352    & 16.65                                                                & \color{red}\textbf{0.913} & \color{red}\textbf{0.853} & \color{red}\textbf{0.895} & \color{red}\textbf{0.005} & 0.778  & 0.705  & 0.763 & \color{red}\textbf{0.031} \\ \hline \hline
\end{tabular}}
\label{input size}
\end{table}

\subsection{\textbf{Ablation Studies}}
We perform a series of experiments to verify the effectiveness of all components, namely the spectrum-guided non-local perception (SNP) module, the multi-source semantic extractor (MSE) and the dense cross-layer interaction (DCI) decoder. In addition, we conduct ablation studies on the hyperparameters ($i.e.$, decoder channels, and input {sizes}). All experiments are conducted using ResNet50 \cite{ResNet}, and performance is evaluated on CVC-300 \cite{CVC-300} and CVC-ColonDB \cite{ColonDB} datasets.

\subsubsection{Effectiveness of SNP module}
{The} spectrum-guided non-local perception (SNP) module is designed to model global long-range dependencies and local details for enhancing the discriminative ability of initial features. As depicted in Table \ref{module study}, the ``Baseline'' (Tab.  \ref{module study} (a)) consists of a ResNet50 \cite{ResNet} encoder and a feature pyramid network (FPN) \cite{FPN} decoder. Subsequently, we integrate the ``SNP'' module (Tab. \ref{module study} (b)) into the ``Baseline'', and it can be seen that {the} accuracy has improved significantly. In particular, performance has increased by {2.93\%, 4.83\%, 4.58\%, and 25\%} on CVC-300 \cite{CVC-300} under four evaluation metrics. Similar boosts can be observed for CVC-ColonDB \cite{ColonDB} dataset. Furthermore, we validate the effectiveness of the spectral information within the SNP module. From Table \ref{module1_ag_c} {(a)}–(c), we observe that removing the spectrum information leads to degraded polyp segmentation performance compared with the complete SNP module in Table \ref{module1_ag_c} (d) on CVC-300 \cite{CVC-300}. In addition, we compare our SNP module with existing transformer-based feature enhancement modules, including the Swin Transformer block (STB, Table \ref{module1_com_c} (a)) \cite{Swin} and the Restormer block (RB, Table \ref{module1_com_c} (b)) \cite{Restormer}. In particular, our SNP module shows clear advantages in the metric $\mathcal{M}$, achieving 0.008 $vs.$ 0.008 and 0.014 on CVC-300 \cite{CVC-300}, and 0.033 $vs.$ 0.039 and 0.039 on CVC-ColonDB \cite{ColonDB}. Figs. \ref{all module}–\ref{visual module1-duibi} show the corresponding visual results, where it can be observed that the polyps segmented using our SNP module are closer to the ground truth.

\subsubsection{Effectiveness of MSE}
The objective of the multi-source semantic extractor (MSE) is to capture higher-level semantic information with different receptive fields to aid in the coarse localization of polyps. From Table \ref{module study}, it can be observed that, compared {with} the ``Baseline'' (Table \ref{module study} (a)), {incorporating} the ``MSE'' (Table \ref{module study} (c)) effectively improves the segmentation accuracy of intestinal polyps on CVC-300 \cite{CVC-300} and CVC-ColonDB \cite{ColonDB} datasets. Subsequently, we validate the compatibility of {our} ``MSE'' and ``SNP'' {components} in Table \ref{module study} (d). {It is evident that, compared with} using the ``MSE'' (Table \ref{module study} (c)) or the ``SNP'' module (Table \ref{module study} (b)) {individually}, the model achieves higher detection accuracy when {the two} are combined. Specifically, on CVC-300 \cite{CVC-300}, performance improved in three evaluation metrics ($i.e.$, $Dic$, $IoU$, and $F_{m}^{w}$) by {2.52\%, 4.05\%, 4.09\% and 2.05\%, 2.49\%, 2.49\%}, respectively. These results provide strong evidence of the compatibility between the ``MSE'' and the ``SNP'' module. Furthermore, Tab. \ref{module2} (a) {corresponds} to the ``MSE'' variant without the adaptive spectrum filter (ASF), which captures spectrum information with an image-level receptive field. Tabs. \ref{module2} (b)–(c) report the performance of classic semantic extractors, $i.e.$, atrous spatial pyramid pooling (ASPP) \cite{ASPP} and the receptive field block (RFB) \cite{RFB}. It can be seen that the designed ``MSE'' has strong competitiveness in Fig. \ref{visual module2-duibi}. In addition, in Tabs. \ref{module2-fr} (a)-(e), we analyze the filling rates of atrous convolutions in the ``MSE'', selecting four different filling rates: (1, 1, 1, 1, 1, 1), (1, 2, 3, 4, 5, 6), and (2, 4, 6, 8, 12, 16). It can be observed that the {best} performance is {achieved} when the filling {rates are} set to (3, 6, 9, 12, 15, 18). These experimental results {demonstrate} that the MSE plays a crucial role in the overall performance of the ASGNet model.

\subsubsection{Effectiveness of DCI decoder}
{In dense} cross-layer interaction (DCI) decoder integrates high-level semantic and low-level spatial information, generating high-quality representations to better segment polyps through a series of complementary optimizations. In Table \ref{module study}, the ``DCI'' decoder (Tab. \ref{module study} (e)) {shows a clear performance improvement} compared to {the FPN-only} structure in the ``Baseline'' (Tab. \ref{module study} (a)). It achieves notable improvements of {1.76\% and 3.53\%} for $Dic$ and $IoU$ {in} CVC-300 \cite{CVC-300}, along with enhancements of {4.94\% and 7.32\%} {in} CVC-ColonDB \cite{ColonDB}. Subsequently, we validate the compatibility of the ``DCI'' decoder with the ``SNP'' module and the ``MSE''. {From Tabs. \ref{module study} (f)–(h),} it can be observed that {integrating} the ``DCI'' decoder into both the ``SNP'' module and the ``MSE'' leads to noticeable performance improvements. In addition, we {analyze} the internal structure of the ``DCI'' decoder, where ``DCI w/o ASF'' refers to the removal of spectrum information and ``DCI w/o Edg.'' denotes the removal of edge information. From Fig. \ref{visual module3} and Table \ref{module3}, {we} observe that the segmentation accuracy of the ``DCI'' decoder {drops markedly when both} spectrum and edge information are removed, confirming that these cues significantly enhance the ability to accurately segment polyps.

\subsubsection{Hyper-parameters analysis}
In Table \ref{channels}, we present an ablation {study} on the decoder channel settings, {evaluating channels} of 32, 64, 96, 128, and 160. As shown in Tables \ref{channels} (a)–(e), ASGNet {achieves} strong performance under different channel settings. Furthermore, we {conduct} a sensitivity analysis on the hyperparameters related to input {image} sizes, as {reported} in Table \ref{input size}. The input image sizes are {set to} 256$\times$256, 288$\times$288, 352$\times$352, 384$\times$384, and 416$\times$416. Rows (a)–(e) in Table \ref{input size} {show that} {our} ASGNet model maintains {excellent} performance across these different resolutions. Considering both the parameters and FLOPs, we ultimately select a decoder channel of 96 and an input resolution of 352$\times$352.

\begin{figure}[]
	\centering\includegraphics[width=0.48\textwidth,height=3cm]{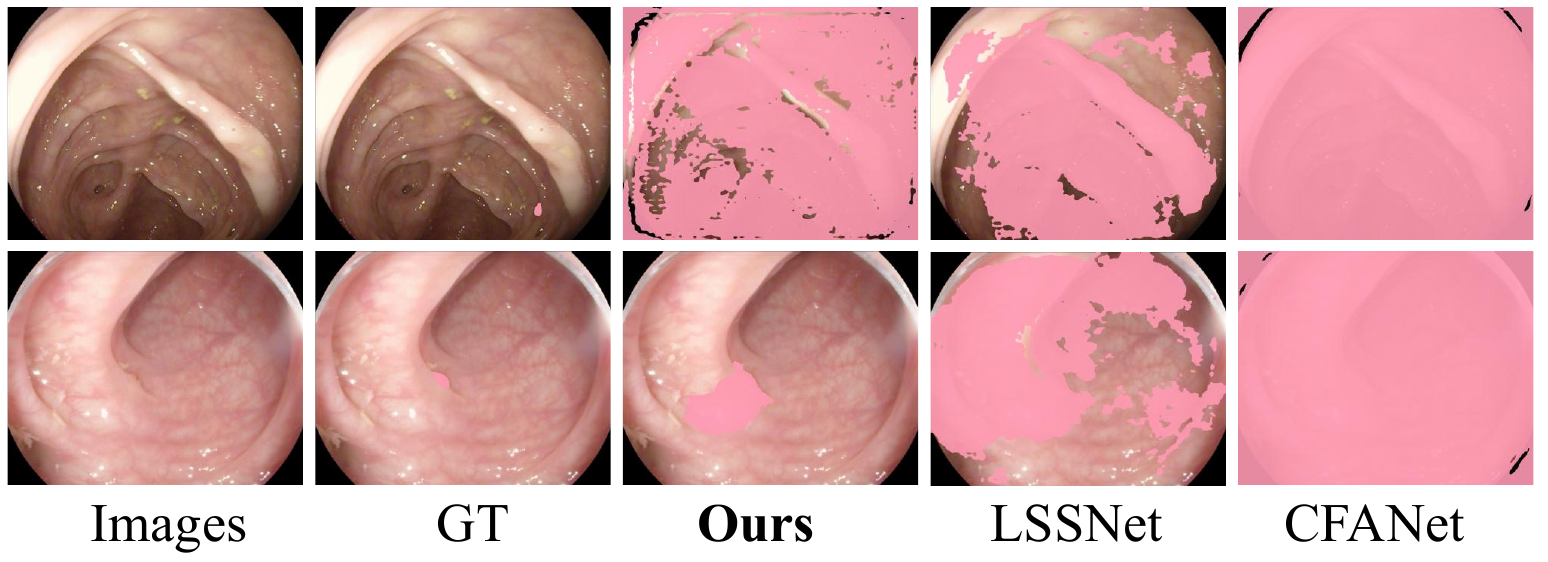}
	\captionsetup{font={small}, justification=raggedright}
	\caption{Some failure cases of our ASGNet method.}
	\label{failure}
\end{figure}

\subsection{Limitations and Analysis}
Although the proposed ASGNet achieves remarkable performance, it still has certain limitations. As illustrated in Fig. \ref{failure}, when dealing with extremely small polyps in complex intestinal environments, ASGNet fails to produce accurate segmentation. To address these challenges, we will explore several potential strategies. First, we plan to increase the representation of extremely small polyps in the training dataset. Second, we aim to deploy larger models to better detect small targets by leveraging their stronger capabilities. These strategies will be key directions in our future work.

\section{Conclusion}
\label{S5}
In this paper, we propose a novel Adaptive Spectrum Guidance Network for automatic polyp segmentation. Specifically, the spectrum-guided non-local perception module is designed to enhance the representational capacity of initial features by integrating local and global information. Meanwhile, a multi-source semantic extractor is introduced to capture high-level semantic information to aid in coase polyp localization. Furthermore, the dense cross-layer interaction decoder is constructed to integrate diverse information. Extensive experiments on multiple datasets demonstrate that our ASGNet method outperforms 21 SOTA models in prediction accuracy.

\ifCLASSOPTIONcaptionsoff
\newpage
\fi

\bibliographystyle{./IEEEtran}
\bibliography{./ASGNet}
\begin{IEEEbiography}
[{\includegraphics[width=1in,height=1.25in,clip,keepaspectratio]{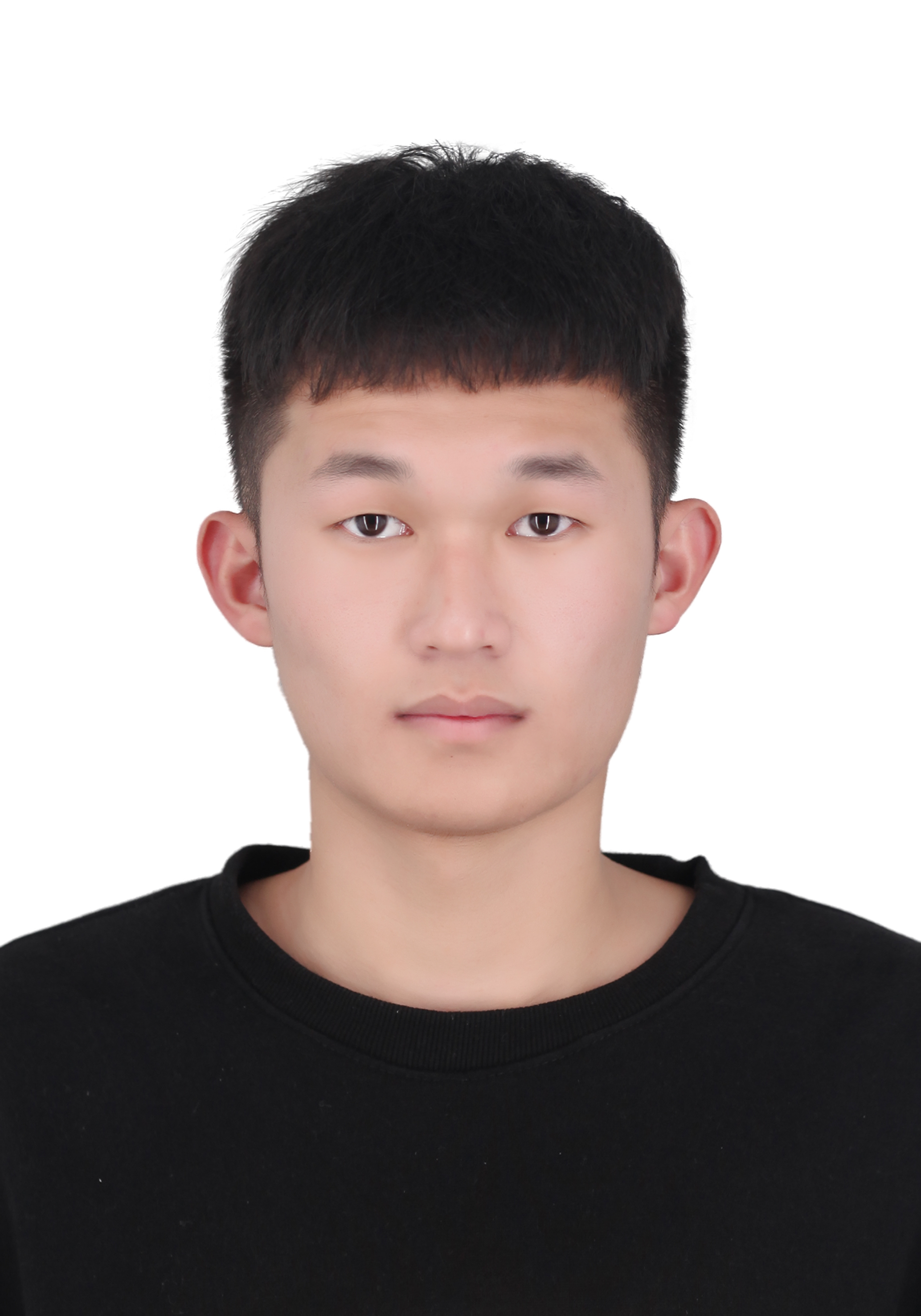}}]{Yanguang Sun} is currently pursuing his Ph.D. at Nanjing University of Science and Technology (NJUST), Nanjing, Jiangsu, China, under the supervision of Professor Lei Luo. His research interests focus on computer vision, and image processing. 

He has published multiple papers as the first author, covering ECCV, ICCV, IJCAI, AAAI, IEEE TNNLS, IEEE TGRS, and IEEE TCSVT.

He has served as a reviewer for prestigious conference/journal, such as CVPR, ECCV, ICCV, AAAI, IJCAI, NeruIPS, IEEE TCSVT, IEEE TMM,  IEEE TNNLS, IEEE TIP, IEEE TCYB and PR.	
\end{IEEEbiography}

\begin{IEEEbiography}
[{\includegraphics[width=1in,height=1.25in,clip,keepaspectratio]{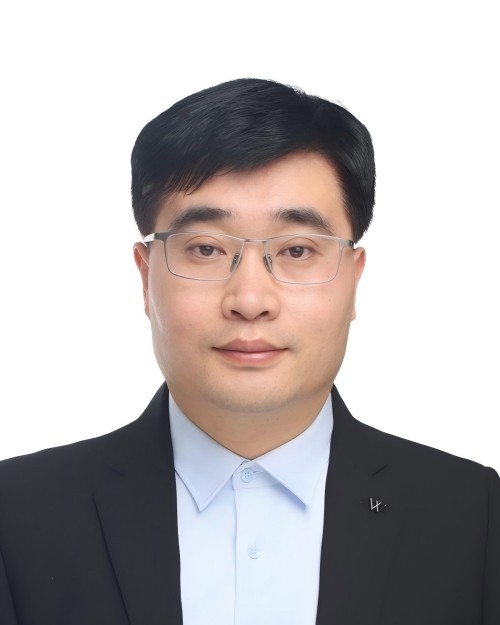}}]{Hengmin Zhang} received the PhD degree in computer science and engineering from the School of Computer Science and Engineering, Nanjing University of Science and Technology, China. He subsequently held postdoctoral research positions at the School of Information Science and Engineering, East China University of Science and Technology, Shanghai, and the Department of Computer and Information Science, University of Macau, followed by a research fellowship at the School of Electrical and Electronic Engineering, Nanyang Technological University, Singapore. He is currently a Professor with the School of Information Science and Engineering, East China University of Science and Technology. He has authored or co-authored more than 30 technical papers in top-tier international journals and conferences. In recognition of his research excellence, he has received the Outstanding Doctoral Dissertation Award from both the Chinese Institute of Electronics (CIE) and Jiangsu Province. His research interests span pattern recognition, intelligent systems, data driven science, and optimization-based learning.

\end{IEEEbiography}

\begin{IEEEbiography}
[{\includegraphics[width=1in,height=1.25in,clip,keepaspectratio]{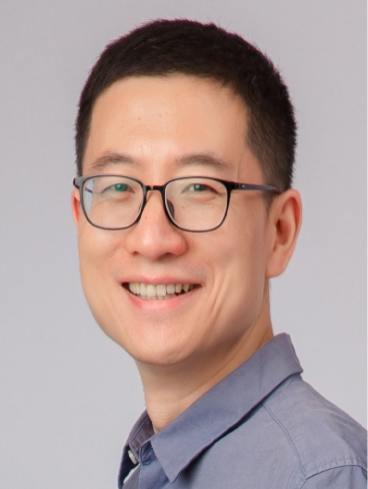}}]{Jianjun Qian} received the Ph.D. degree in pattern recognition and intelligence systems from Nanjing University of Science and Technology (NJUST), Nanjing, China, in 2014. 

He is currently a Processor with NJUST. His research interests include pattern recognition theory, computer vision, and machine learning. 

Prof. Qian was selected as a Hong Kong Scholar in 2018. He has served as the Guest Editor for Neural Processing Letters and The Visual Computer	.
\end{IEEEbiography}

\begin{IEEEbiography}
[{\includegraphics[width=1in,height=1.25in,clip,keepaspectratio]{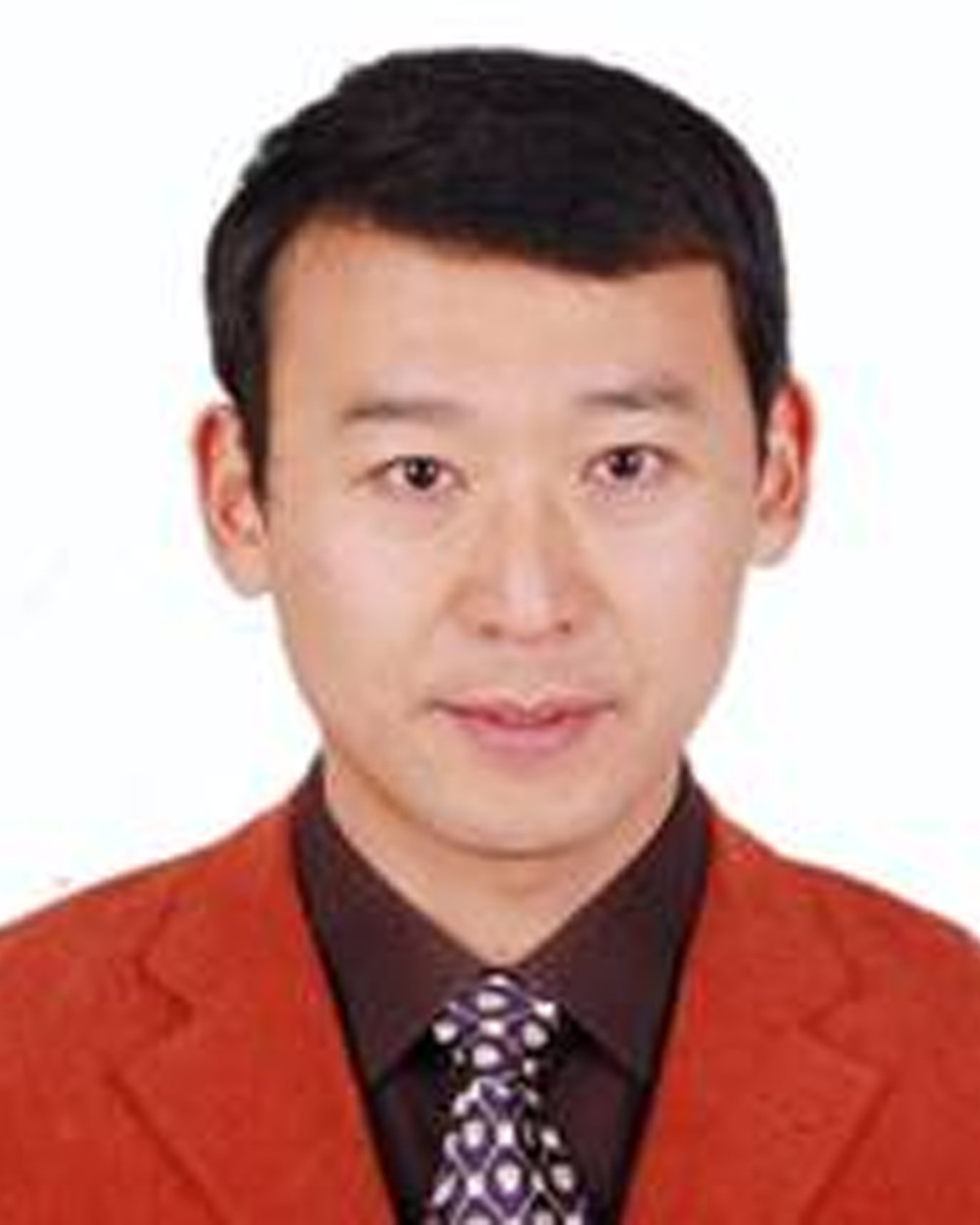}}]{Jian Yang} received the PhD degree from Nanjing University of Science and Technology (NJUST) in 2002, majoring in pattern recognition and intelligence systems. 
From 2003 to 2007, he was a Postdoctoral Fellow at the University of Zaragoza, Hong Kong Polytechnic University and New Jersey Institute of Technology, respectively. From 2007 to present, he is a professor in the School of Computer Science and Technology of NJUST. 

He is the author of more than 300 scientific papers in pattern recognition and computer vision. His papers have been cited over 60000 times in the Scholar Google. His research interests include pattern recognition and computer vision. 

Prof. Yang is a fellow of International Association for Pattern Recognition (IAPR). He is/was an associate editor of Pattern Recognition, Pattern Recognition Letters, IEEE Trans. Neural Networks and Learning Systems, and Neurocomputing.
\end{IEEEbiography}

\begin{IEEEbiography}
[{\includegraphics[width=1in,height=1.25in,clip,keepaspectratio]{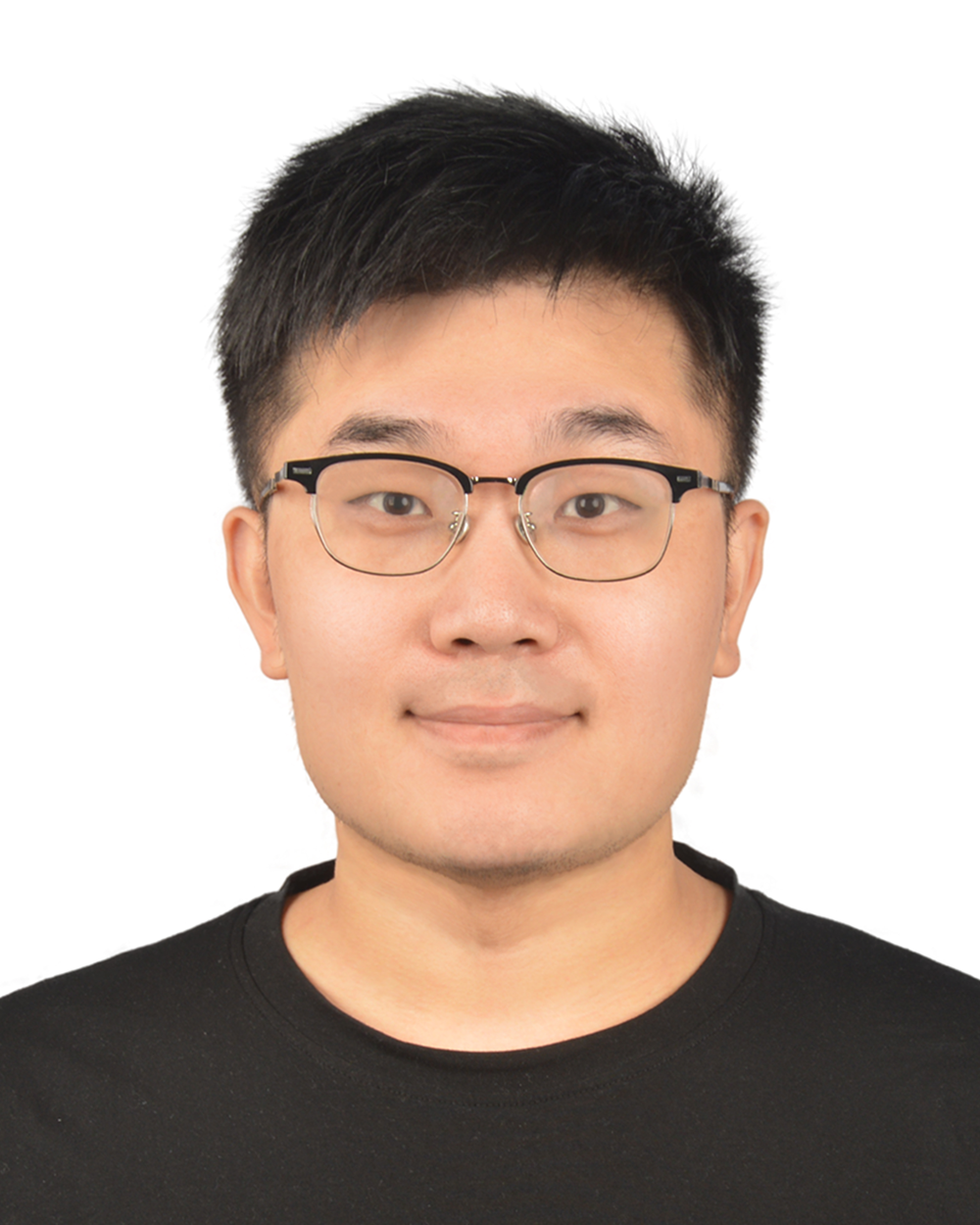}}]{Lei Luo} received the Ph.D. degree in pattern recognition and intelligence systems from the School of Computer Science and Engineering, Nanjing University of Science and Technology (NJUST), in 2016. From 2017 to 2020, he was a Post-Doctoral Fellow at the University of Texas at Arlington, TX, USA, and the University of Pittsburgh, PA, USA. 

He is currently a Professor in the School of Computer Science and Technology of NJUST. His research interests include pattern recognition, machine learning, data mining and computer vision. 

Prof. Luo has served as an PC/SPC Member for IJCAI, AAAI, NeurIPS, ICML, KDD, CVPR, and ECCV, and a reviewer for over ten international journals, such as IEEE TPAMI, IEEE TIP, IEEE TSP, IEEE TCSVT, IEEE TNNLS, IEEE TKDE, and PR.	
\end{IEEEbiography}

\end{document}